\documentclass[lettersize,journal]{IEEEtran}
\usepackage{amsmath,amsfonts}
\usepackage{algorithmic}
\usepackage{algorithm}
\usepackage{array}
\usepackage{textcomp}
\usepackage{stfloats}
\usepackage{url}
\usepackage{verbatim}
\usepackage{graphicx}
\usepackage{cite}
\usepackage{hyperref}
\usepackage{url}
\usepackage{threeparttable}
\usepackage{amssymb}
\usepackage{mathtools}
\usepackage{amsthm}

\usepackage{booktabs}       
\usepackage{subcaption}    
\usepackage{enumitem}
\usepackage{wrapfig}

\usepackage[table]{xcolor}
\usepackage{notation}
\usepackage{xspace}
\usepackage{bbding}
\usepackage{etoc}

\usepackage[capitalize,noabbrev]{cleveref}

\newtheorem{thm}{Theorem}

\etocdepthtag.toc{mtchapter}
\etocsettagdepth{mtchapter}{subsubsection}
\etocsettagdepth{mtappendix}{none}

\def\revise{\textcolor{black}}

\newcommand{\sexyname}{TCA-Attention\xspace}
\newcommand{\sexynamellm}{TCA-Attention\xspace}
\newcommand{\methodname}{Training-free Context-adaptive Attention\xspace}
\def\ie{\mbox{\textit{i.e.}}}

\def\mytitle{Training-free Context-adaptive Attention for Efficient Long Context Modeling}

\begin{document}

\title{\mytitle}

\author{Zeng You, Yaofo Chen, Shuhai Zhang, Zhijie Qiu, Tingyu Wu, Yingjian Li, Yaowei Wang, Mingkui Tan

\thanks{Zeng You is with the School of Future Technology, South China University of Technology, Guangzhou, China and also with Pengcheng Laboratory, Shenzhen, China (e-mail: zengyou.yz@gmail.com).}
\thanks{Yaofo Chen is with the School of Future Technology, South China University of Technology, Guangzhou, China (e-mail: chenyaofo@gmail.com).}
\thanks{Shuhai Zhang, Zhijie Qiu, and Tingyu Wu are with the School of Software Engineering, South China University of Technology, Guangzhou, China (e-mail: shuhaizhangshz@gmail.com, 202420145073@mail.scut.edu.cn, 202230490150@mail.scut.edu.cn.}
\thanks{Yingjian Li is with Pengcheng Laboratory, Shenzhen, China (e-mail: yingjian1122@gmail.com).}
\thanks{Yaowei Wang is with Harbin Institute of Technology, Shenzhen, China, and also with  Pengcheng Laboratory, Shenzhen, China (e-mail: wangyaowei@hit.edu.cn).}
\thanks{Mingkui Tan is with the School of Software Engineering, South China University of Technology, Guangzhou 510006, China (e-mail: mingkuitan@scut.edu.cn).}
\thanks{Zeng You, Yaofo Chen, and Shuhai Zhang are equal contributors.}
\thanks{Mingkui Tan and Yaowei Wang are corresponding authors.}
}

\markboth{Journal of \LaTeX\ Class Files,~Vol.~14, No.~8, August~2021}%
{Shell \MakeLowercase{\textit{et al.}}: A Sample Article Using IEEEtran.cls for IEEE Journals}

\IEEEpubid{0000--0000/00\$00.00~\copyright~2021 IEEE}

\maketitle

\begin{abstract}
Large Language Models (LLMs) have demonstrated remarkable capabilities across a wide range of natural language processing tasks. These capabilities stem primarily from the self-attention mechanism, which enables modeling of long-range dependencies. However, the quadratic complexity of self-attention with respect to sequence length poses significant computational and memory challenges, especially as sequence length extends to extremes. While various sparse attention and KV cache compression methods have been proposed to improve efficiency, they often suffer from limitations such as reliance on fixed patterns, inability to handle both prefilling and decoding stages, or the requirement for additional training. In this paper, we propose \methodname (\sexyname), a training-free sparse attention mechanism  that selectively attends to only the informative tokens for efficient long-context inference. Our method consists of two lightweight phases: i) an offline calibration phase that determines head-specific sparsity budgets via a single forward pass, and ii) an online token selection phase that adaptively retains core context tokens using a lightweight redundancy metric. \sexyname provides a unified solution that accelerates both prefilling and decoding while reducing KV cache memory footprint, without requiring parameter updates or architectural changes. Theoretical analysis shows that our approach maintains bounded approximation error. Extensive experiments demonstrate that \sexyname achieves a \textbf{2.8$\times$} speedup and reduces KV cache by \textbf{61\%} at 128K context length while maintaining performance comparable to full attention across various benchmarks, offering a practical plug-and-play solution for efficient long-context inference.
\end{abstract}


\section{Introduction}\label{sec:intro}
\IEEEPARstart{L}{arge} language models (LLMs) have revolutionized numerous natural language processing tasks, establishing themselves as indispensable tools for modern AI applications~\cite{de2024human,lin2024correctable,zhang2024vision}. Models such as the LLaMA series~\cite{touvron2023llama2,meta2024llama3}, GPT-family models~\cite{brown2020gpt3,openai2023gpt4}, and recent breakthroughs like GPT-o1~\cite{openai2024openaio1card} and DeepSeek-R1~\cite{guo2025deepseek} have demonstrated exceptional capabilities in complex tasks, including multi-step reasoning~\cite{wei2022chain,Maciej2025Demystifying} and document-level comprehension~\cite{pasunuru2021data,zhang2025TDGI}. The remarkable performance of these models predominantly stems from the self-attention mechanism~\cite{vaswani2017attention}, which enables comprehensive dependency modeling across entire input sequences through dense pairwise token interactions.

However, this expressive power comes at a substantial computational cost, posing a significant challenge for efficient long context modeling. As context lengths extend to extremes (e.g., 128K tokens), the quadratic complexity of self-attention presents severe bottlenecks in both computation and memory. For a sequence of length $L$, the prefilling stage, which computes attention across all token pairs, exhibits $\mathcal{O}(L^2)$ computational complexity, while the decoding stage suffers from $\mathcal{O}(L)$ memory access overhead per generated token due to the linearly growing key-value (KV) caches. More critically, not all tokens contribute equally to the final output. The presence of redundant or minimally informative tokens dilutes attention allocation to semantically critical content, thereby degrading both efficiency and accuracy~\cite{chen2025core,zhang2025dga}. This challenge becomes increasingly pronounced in applications involving lengthy documents, extended conversations, or complex multi-modal contexts.

\IEEEpubidadjcol
To mitigate these issues, existing approaches have pursued several distinct directions. Early efforts, such as static sparse attention~\cite{zaheer2020big,beltagy2020longformer,xiao2024efficient}, employ predefined attention patterns to reduce computation, yet they lack adaptability to input content and often compromise performance on context-sensitive tasks. More recent prefilling-stage dynamic methods~\cite{jiangminference,laiflexprefill,xu2025xattention} adjust attention patterns per head during prefilling, but they are restricted to a limited set of handcrafted sparsity patterns and do not address inefficiency in the decoding phase. On the other hand, decoding-stage KV cache compression techniques~\cite{li2024snapkv,qincake,liu2025chunkkv} reduce memory usage by evicting or merging cached KV entries, yet they fail to accelerate prefilling and typically apply uniform compression strategies across attention heads, ignoring the intrinsic diversity of redundancy across heads. While a few unified frameworks, such as DuoAttention~\cite{xiaoduoattention} and Lserve~\cite{yanglserve}, target both prefilling and decoding, they require continued training to determine head-specific patterns and only enable context-adaptive sparsity during decoding.

\begin{figure*}[t]
    \centering
    \includegraphics[width=0.95\linewidth]{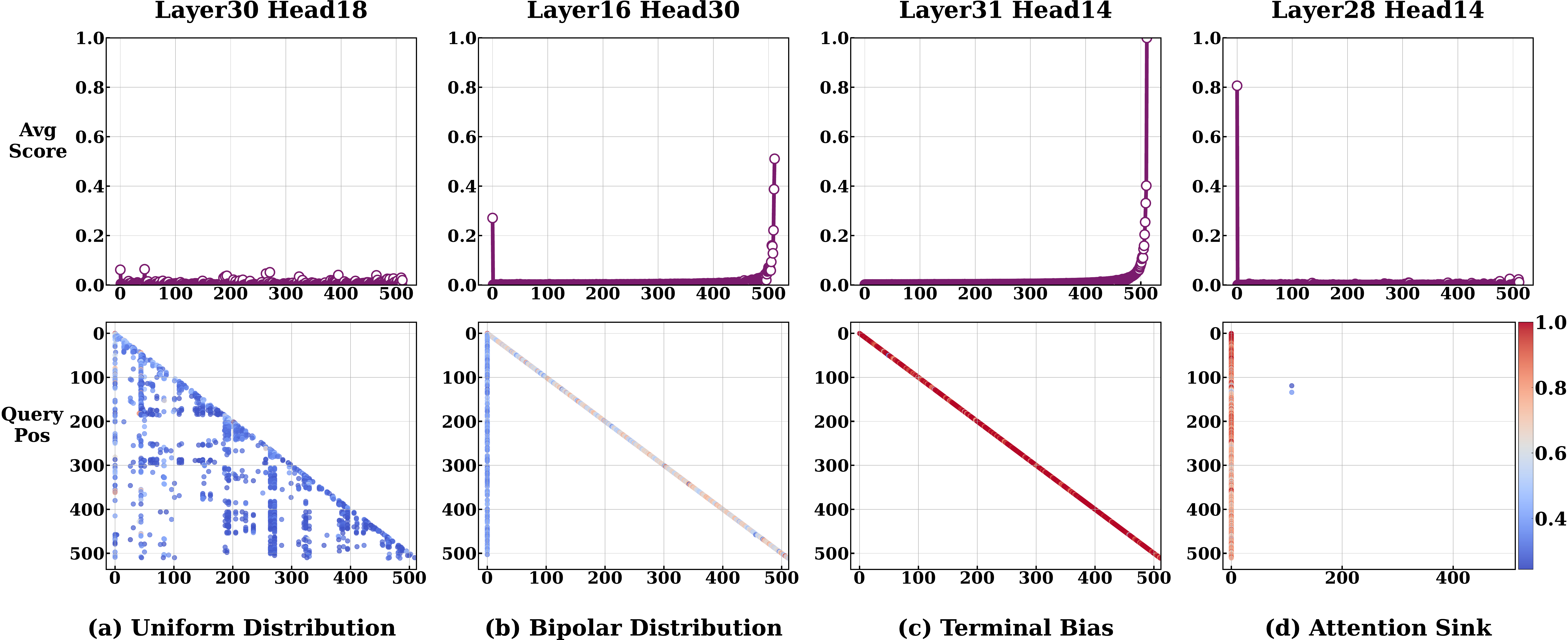}
    \caption{Visualizations of attention distributions in LLaMA-3.1-8B-Instruct: 1) average attention scores across key positions on the first row, and 2) scatter points highlighting attention scores above row-wise averages on the second row. The observations motivate the design principles of \sexyname (refer to Appendix B for more visualizations).}
    \label{fig:motivation}
\end{figure*}

The fundamental limitations of existing literature can be distilled into two critical research gaps. First, despite growing interest in efficient attention mechanisms, most existing methods fail to provide a full \textit{training-free} solution that dynamically accelerates both prefilling and decoding while simultaneously reducing KV cache footprint, without requiring architectural modifications or parameter updates.
Second, they fail to model the intricate nature of attention redundancy adequately. As demonstrated in Figure~\ref{fig:motivation}, different attention heads exhibit diverse redundancy distributions across models, ranging from uniform distributions to attention sinks, aligning with findings in prior works~\cite{beltagy2020longformer, xiao2024efficient, jiangminference}. Meanwhile, token importance varies dynamically within input content. However, existing methods either use uniform compression strategies or rely on a limited set of hand-designed patterns, lacking the fine-grained, context-adaptive mechanism required to handle such head-specific and context-dependent redundancy.

To address these limitations, we propose \methodname (\sexyname), a sparse attention mechanism designed for efficient long context modeling across prefilling and decoding stages. In contrast to prior unified approaches~\cite{xiaoduoattention,yanglserve,xiao2024efficient}, our method inherently requires no parameter updates, architectural adjustments, or reliance on predefined sparse patterns. It operates via two lightweight, complementary phases:
1) An \textit{offline sparsity configuration phase} that estimates head-specific redundancy levels through a single forward pass, determining individualized sparsity budgets per head;
2) An \textit{online core-context selection phase} that dynamically selects a compact set of core tokens per head using a lightweight, context-aware redundancy metric during inference.
This design ensures that \sexyname is both head-aware and input-adaptive, enabling fine-grained sparsity tuning without compromising representational capacity. Moreover, we provide a theoretical guarantee demonstrating that the approximation error of \sexyname is bounded and explicitly controllable.
Our contributions are summarized as:

\begin{itemize}[leftmargin=*]
\item 
Our proposed \sexyname unifies acceleration for prefilling, decoding, and KV cache reduction via dynamically attending to only the informative tokens in the context. Unlike prior approaches relying on predefined patters or retraining, our method eliminates any training overhead and dynamically adapts sparsity to input content, enabling plug-and-play deployment for efficient long-context inference.

\item Our proposed \sexyname integrates two phases that explicitly adapt to both head-specific and context-dependent attention redundancy: 1) an offline calibration phase for head-specific sparsity configuration, and 2) an online token selection phase with a lightweight, context-aware redundancy metric. This enables fine-grained acceleration while preserving long context modeling performance.

\item We prove that the approximation error introduced by core context selection is explicitly bounded, ensuring reliability without training. This principled design is validated empirically. Extensive experiments demonstrate that our method achieves a \textbf{2.8$\times$} speedup and reduces KV cache by \textbf{61\%} at 128K context length while maintaining performance comparable to full attention across various tasks.

\end{itemize}

\section{Related Works}
\label{sec:related_works}
\textbf{Efficient Attention}.
Efficient attention mechanisms address the quadratic complexity of self-attention and KV cache memory bottlenecks through several major strategies. Early Static sparse attention mechanisms~\cite{zaheer2020big,beltagy2020longformer,xiao2024efficient} reduce computational complexity by imposing predefined sparsity patterns on attention matrices. While effective in certain long-document tasks and easy to deploy, these approaches lack adaptivity. Another line of work~\cite{katharopoulos2020transformers,sun2023retentive} 
reformulates self-attention as a linear dot-product of kernel feature maps and leverages the associativity property of matrix products to achieve linear complexity. These methods reduce theoretical complexity and memory usage, but often introduce accuracy degradation. A complementary direction improves the efficiency of exact attention through IO-aware kernel optimizations, tiling strategies, and parallel execution primitives~\cite{dao2022flashattention,dao2024flashattention, hong2024flashdecoding++,wang2025enabling}. Such implementations significantly boost runtime performance and memory efficiency for full attention. Our method is orthogonal to this direction.

More recent work~\cite{jiangminference,laiflexprefill,xiaoduoattention} proposes dynamic or training-free sparse attention mechanisms that target the prefilling stage, where the entire input sequence is processed at once. These methods typically select patterns from predefined sets at the head level. While they achieve acceleration during prefilling, they do not reduce KV-cache growth and thus provide limited benefits for autoregressive decoding. During generation, the primary bottleneck is the linear growth of the KV-cache. During generation, the primary bottleneck is the linear growth of the KV-cache. KV cache compression techniques reduce computation and KV cache during generation through eviction~\cite{li2024snapkv,qincake,liu2025chunkkv} or merging~\cite{wan2024d2o} tokens, yet cannot accelerate prefilling and often apply uniform head-wise compression. Recent studies~\cite{chen2025core, yuan-2025-native, lu2025moba, zhang2025dga} like DuoAttention~\cite{xiaoduoattention} and Lserve~\cite{yanglserve} attempt to bridge both stages using consistent sparsity patterns. However, they often require additional model training, complex profiling, or intrusive system-level changes.

In contrast to the above categories, our \sexyname provides a training-free, head-aware, and context-adaptive sparse attention mechanism that simultaneously accelerates prefilling and decoding. It reduces computation and KV-cache size within a single unified framework, while offering explicit theoretical error guarantees and requiring no model retraining, architecture modification, or heavy offline tuning. This distinguishes our approach both methodologically and in practical deployability.

\textbf{Long-Context Language Modeling}.
Handling long input sequences is critical for applications requiring document-level reasoning, code generation, or extended dialogue history\cite{liu2024lost,bai2023longbench,hsieh2024ruler}. 
On the data front, tailored curation and synthesis strategies are developed to equip models with long-range dependency awareness. For the pre-training stage, the methods focus on filtering for long-context quality via linguistic metrics or attention-pattern analysis, mixing domains to preserve diversity while upsampling long sequences, and synthesizing data by clustering related passages or constructing ``knotted'' documents to enhance structural reasoning~\cite{liu2024longwanjuan,fu2024data}, During the post-training stage, instruction data are synthesized to mitigate positional bias and to support multi-hop reasoning over dispersed information~\cite{liu2024lost,an2024make}. 

Architecturally, Rotary Position Embedding (RoPE)~\cite{su2024roformer} and its variants serve as the foundation; extrapolation techniques such as NTK-aware scaling~\cite{peng2023ntk}, and YaRN~\cite{peng2024yarn} systematically map out-of-distribution positions into the training range, often without fine-tuning. Beyond embeddings,  significant direction replaces quadratic attention with linear-complexity models, such as selective state-space models (Mamba)~\cite{gu2024mamba}, the RetNet architecture with recurrent and parallel paradigms~\cite{roy2023retnet}, and Test-Time Training that dynamically stores context in fast weights~\cite{sunlearning}. These efforts enable models to scale to longer contexts. Our work aligns with this goal by introducing a training-free, adaptive sparse attention mechanism that accelerates both prefilling and decoding.

\section{Motivations}
\label{sec: motivations}

\subsection{Understanding Bottlenecks of Self-Attention in Long-context Modeling}
Most existing LLMs are built on the Transformer~\cite{vaswani2017attention} architecture, where the self-attention mechanism serves as the core module for capturing global contextual dependencies. Given an input sequence $\bX  = \{\bx_1, \bx_2, \ldots, \bx_L\}\in \mathbb{R}^{L \times d}$ of $L$ tokens with model dimension $d$, The multi-head self-attention mechanism computes contextualized representations through $h$ heads:
\begin{equation}
\begin{split}
    & \text{MultiHead}(\bX) = [{\bf{Att}}^1, \ldots, {\bf{Att}}^h] \in \mathbb{R}^{L \times d}, \\
    &{\bf{Att}}^i = \text{softmax}\left(\frac{\bQ^i{\bK^i}^\top}{\sqrt{d_h}}\right)\bV_i, \\
    &\bQ^i=\bX\bW^{Q_i},~ \bK^i=\bX\bW^{K_i},~\bV_i=\bX\bW^{V_i},
\end{split}
\end{equation}
where $d_h$ is the head dimension (typically $d_h = d/h$), $\bW^{Q_i}, \bW^{K_i}, \bW^{V_i}\in \mathbb{R}^{d\times d_h}$ are learnable parameters for the $i$-th head. 
The multi-head mechanism enables parallel attention operations across distinct feature subspaces, facilitating position-aware information aggregation.

\textbf{Challenges of Self-Attention in Handling Context Redundancy}: 
As the context length $L$ grows, the context inevitably exhibits redundant information~\cite{jiangminference,chen2025core, zhang2025dga}. Vanilla self-attention faces three challenges in handling such redundancy: 
1) \textbf{Quadratic computational complexity}: It incurs $O(L^2)$ computational cost by computing pairwise attention scores across all tokens. This leads to excessive computation when much of the context is redundant.
2) \textbf{Memory growth in key-value (KV) Cache}: The KV cache, which grows linearly with $L$, presents a major deployment bottleneck. For instance, processing a 128K sequence with LLaMA2-7B requires 64GB GPU memory for KV cache, exceeding the capacity of most GPUs.
3) \textbf{Interference from irrelevant tokens}: More critically, irrelevant tokens degrade the model’s ability to focus on critical information, thereby harming performance.

\subsection{Exploration and Exploitation of Attention Redundancy Properties}\label{sec:pattern}

The severe inefficiency of self-attention in long contexts raises a critical question: \textit{given the aforementioned head-specific and context-dependent nature of attention redundancy, how can we design a unified framework that exploits these properties to accelerate both prefilling and decoding?} As visualized in Figure~\ref{fig:motivation} and supported by prior works~\cite{jiangminference,xu2025xattention,laiflexprefill,xiao2024efficient}, attention redundancy manifests heterogeneously across heads and varies dynamically with input context. The fundamental challenge, however, lies not in observing these properties but in addressing them simultaneously and effectively within a single, training-free acceleration framework.

Existing approaches only address a subset of this challenge. \textit{Static sparse methods}~\cite{beltagy2020longformer, zaheer2020big, xiao2024efficient} ignore both dynamic context and head specificity. \textit{Prefilling-only dynamic attention} methods~\cite{jiangminference,laiflexprefill} adapt per head but are constrained to a small set of hand-designed patterns and, critically, fail to accelerate decoding. \textit{Decoding-only KV compression} methods~\cite{li2024snapkv, qincake, wan2024d2o} reduce memory but cannot accelerate prefilling and often apply uniform policies across heads within a layer. While recent unified methods like DuoAttention~\cite{xiaoduoattention} and Lserve~\cite{yanglserve} target both stages, they require continuous training for pattern determination and operate with fixed sparse patterns, lacking context-aware adaptability.

\textbf{Design Principles for Efficient Attention}. This limitation reveals a fundamental design gap: no existing method simultaneously achieves dynamic context adaptation, head-aware sparsity, and training-free deployment across both prefilling and decoding stages. We therefore distill the requirements for an ideal solution into four key principles:

\begin{enumerate}[left=0pt]
    \item \textbf{Sparse Computation}:
    Selectively attend to a small subset of critical tokens and discard irrelevant ones due to the tokens' redundancy in both prefilling and decoding stages.
    \item \textbf{Dynamic Adaptation}: The sparsity must be input-dependent, not predetermined by fixed sparse patterns.
    \item \textbf{Head-aware Sparsity Customization}: Sparsity strategies must be tailored to the redundancy level of each head.
    \item \textbf{Training-free Deployment}:
    Eliminates the need for parameter updates, architectural modifications, or retraining, enabling immediate plug-and-play deployment without compromising model performance.
\end{enumerate}
Existing methods fulfill only a subset of these principles, creating a performance-efficiency gap. Static methods violate (2) and (3); prefilling-only and decoding-only methods violate (1); existing unified approaches violate (2) and (4). This motivates our design of a unified attention mechanism that satisfies all four principles simultaneously, \ie, sparsity, dynamic, head-aware customization, and training-free.

\begin{figure*}[t]
    \centering
    \includegraphics[width=0.95\linewidth]{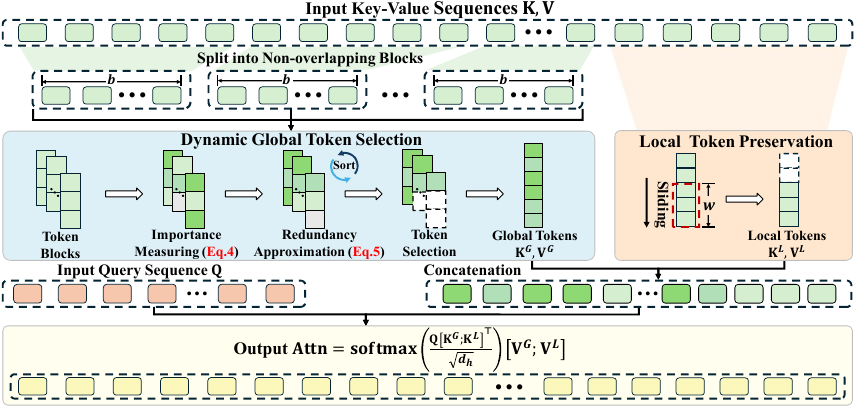}
    \caption{Overview of our \sexyname. We dynamically select a subset of key/value tokens, which combines 1) a global subset $\bK^G,\bV^G$, selected online based on a pre-determined configuration (Sec.~\ref{sec:online}), to model long‐distance dependencies; and 2) a local subset $\bK^L,\bV^L$, preserving neighboring tokens to capture fine‐grained local context. The concatenation of these complementary subsets is used for the final attention computation in Eq.~(\ref{eq:atten}).}
    \label{fig:method}
\end{figure*}

\section{\methodname}
\label{sec:method}

In this paper, we propose \methodname (\sexyname) that simultaneously accelerates prefilling, reduces decoding latency, and compresses KV cache by dynamically selecting core tokens in a head-specific and context-aware manner.
Our method is completely training-free, which ensures seamless integration into existing LLMs.

\textbf{Overview of \methodname.} Our \sexyname includes two phases: 1) During the offline sparsity configuration determination phase (Section~\ref{sec:offline}), we adopt a small calibration dataset to estimate the redundancy level of each head and determine its appropriate sparsity configuration (i.e., the number of tokens to preserve). This phase is performed only once per model.
2) During inference, we dynamically select active tokens in each head based on the determined sparsity configuration (Section~\ref{sec:online}). Theoretically, we prove that the approximation error of \sexyname is bounded and controllable (Section~\ref{sec:theorem}).
We implement our method using Triton to enable efficient parallelization across attention heads. We illustrate the overview in Figure \ref{fig:method} and present the pseudo-code in Algorithms~\ref{alg:offline_search}~and~\ref{alg:dynamic_attention}.

To capture both long-range and short-range dependencies, our dynamic token selection operates on two complementary components: \textit{a global subset} and \textit{a local subset}.
We dynamically select the global subset $\bf K^\text{G}={\bf K}_{\mathcal{S}}$ and $\bf V^\text{G}={\bf V}_{\mathcal{S}}$ from the entire context based on the head-specific sparsity configuration, where $ \mathcal{S} \subseteq \{1, \dots, L\} $ is the selected token index set. We detail the computation of $\mathcal{S}$ in Section~\ref{sec:online}. This subset is responsible for modeling long-distance dependencies.
On the other hand, we always preserve the most recent $w$ tokens for each query as the local subset $\bf K^\text{L}$ and $\bf V^\text{L}$ to capture fine-grained local context, a critical element highlighted by prior works~\cite{zong2021Long, yang2021context, xiao2024efficient} and our analysis. Notably, we ensure no overlap between the global and local subsets to avoid duplicated computation.
Given the query matrix ${\bf Q} \in \mathbb{R}^{L \times d_h}$, our \sexyname computes the attention output as follows:
\begin{equation}\label{eq:atten}
{\bf Att} = \text{Softmax}\left( \frac{{\bf Q} [{\bf K}^\text{G}; {\bf K}^\text{L}]^\top}{\sqrt{d_h}} \right) [{\bf V}^\text{G}; {\bf V}^\text{L}].
\end{equation}
Since the resulting output $ {\bf Att} \small{\in} \mathbb{R}^{L \times d_h} $ preserves the original sequence length $L$, our method seamlessly integrates into existing LLM architectures without requiring any structural modifications. For clarity and brevity, we describe all operations with respect to a single attention head in the following. We apply the same mechanism independently and in parallel to all heads in all layers.

\begin{algorithm}[t]
  \caption{Offline Head-Specific Sparsity Determination}
  \label{alg:offline_search}
  \begin{algorithmic}[1]\small
    \REQUIRE
    Queries $\bQ \small{\in} \mathbb{R}^{L \times d_h}$,
    keys $\bK \small{\in} \mathbb{R}^{L \times d_h}$, configuration candidates $\mathcal{C} = \{\bp^1, \dots, \bp^M\}$
    threshold $\tau \in [0,1]$.
\STATE Compute attention $\bA \small{=} \text{softmax}\left({\bQ\bK^\top}/{\sqrt{d_h}}\right)$
\STATE Initialize valid set $\mathcal{C}_{\text{valid}} \gets \emptyset$
\FOR{each configuration $\bp^i \in \mathcal{C}$}
    \STATE Compute the selected token indexes $\mathcal{S}_i$ based on $\bp^i$ as described in Section~\ref{sec:online}
    \STATE Compute aggregated score $\ba_i$ via Eq.~(\ref{eq:score})
    \STATE \textbf{if} {$\ba_i \geq \tau$} \textbf{then} $\mathcal{C}_{\text{valid}} \gets \mathcal{C}_{\text{valid}} \cup \{\bp^i\}$
\ENDFOR
\STATE $\bp^* = \mathop{\text{argmin}}_{\bp^i \in \mathcal{C}_{\text{valid}}} |\mathcal{S}_i|$
\ENSURE Configuration $\bp^*$
  \end{algorithmic}
\end{algorithm}

\subsection{Offline Head-Specific Sparsity Determination}\label{sec:offline}
Motivated by the head-specific redundancy property observed in Figure~\ref{fig:motivation} and prior works~\cite{zaheer2020big, xiao2024efficient, jiangminference}, we introduce a one-time offline calibration to determine a unique sparsity configuration for each attention head. This configuration dictates how many tokens (token budget) the head should preserve during inference based on its inherent redundancy level. 
To handle a sequence ${\bf X}$ with the arbitrary length $L$, we operate on a block-wise basis. 
We partition the input sequence $\mathbf{X}$ into $m=\lfloor L/b \rfloor$ non-overlapping blocks of size $b$. Let $\mathcal{K}=\{1,2,4,\dots,b\}$ denote the discrete set of allowable per-block retain counts. 
For each head, we seek a configuration ${\bf p} = [p_k | k \in \mathcal{K}]$, where $\sum_{k \in \mathcal{K}}{p_k}\small{=}1$. Each $p_k \in \mathbf{p}$ represents the proportion of blocks assigned a budget of $k$ tokens, enabling an adaptive policy: the actual number of tokens preserved per block can vary based on its content, while the overall distribution is governed by $\mathbf{p}$.
\textbf{Gaussian-Sampling Configuration Candidates Generation}. 
Given the diverse redundancy levels across heads, we need a set of candidate configurations that smoothly transition from high-efficiency to high-accuracy modes. Creating such a diverse configuration set manually is infeasible, as it would require tuning dozens of hyperparameters across all heads. To solve this, we propose a \textit{log-Gaussian sampling strategy} that generates candidate configurations $\mathcal{C} = \{\mathbf{p}^1, \mathbf{p}^2, \ldots, \mathbf{p}^M\}$ from efficiency-focused to accuracy-focused, controlled by just two intuitive hyperparameters.
Specifically, We model the probability of retaining $k$ tokens using a log-Gaussian distribution centered around $\mu$:
\begin{equation}\label{eq:gaussian}
    p_k = \frac{\Phi[\log_2(k)]}{\sum_{k'\in\mathcal{K}}\Phi[\log_2(k')]}, ~\mathrm{where}~\Phi[x]= \exp \left(-\frac{(x-\mu)^2}{2\sigma^2}\right),
\end{equation}
where $\mu$ controls the center of the token budget in log-space (a larger $\mu$ prioritizes performance via more token requirement) and $\sigma$ regulates diversity in sampled configurations (a larger $\sigma$ explores extreme token budget).
In practice, we generate $M$ configuration candidates by uniformly sweeping $\mu$ from $0$ to $\log_2(b)$ with a fixed $\sigma$: $\mathcal{C}\small{=}\{{\bf p}^1,{\bf p}^2,\ldots,{\bf p}^M\}$, where the $i$-th configuration ${\bf p}^i$ is computed with $\mu^{i} \small{=} \log_2\left(1+\frac{(i-1)\cdot (b-1)}{M - 1}\right)$.
This ensures smooth interpolation between efficiency-oriented and accuracy-oriented sparsity patterns.

\textbf{Cumulative-Score Configuration Determination.}
The goal of offline determination is to find the most efficient (sparsest) configuration from the candidate set $\mathcal{C}$ that maintains the head's performance. We measure performance via the aggregated attention score retained by the selected tokens. For a given candidate $\mathbf{p}^i$, we simulate its token selection on a calibration dataset (using our online mechanism from Section~\ref{sec:online}) to obtain the set of indices $\mathcal{S}_i$. We then compute the sum of the average attention scores for each selected token:
\begin{equation}\label{eq:score}
    \ba_i = \sum_{k \in \mathcal{S}_i} \frac{1}{\|\bA_{:,k}\|_0} \sum_{j=1}^{\|\bA_{:,k}\|_0} \bA_{j,k},
\end{equation}
where ${\bf A} = \text{softmax}({{\bf Q}{\bf K}^\top}/{\sqrt{d_h}}) \in \mathbb{R}^{L \times L}$ is computed on a calibration dataset, $\|{\bf A}_{:,k}\|_0$ denotes the number of non-zero elements in the $k$-th column of $\bf A$.
We select the configuration $\bp^* = \mathop{\text{argmin}}_{\bp^i \in \mathcal{C}_{\text{valid}}} |\mathcal{S}_i|$, where $\mathcal{C}_{\text{valid}}$ denotes a configuration set with each $\bp^i$ satisfying ${\bf a}_i \geq \tau$ and $|\mathcal{S}_i|$ denotes the number of preserved tokens. This selects the configuration that preserves the fewest tokens while retaining at least a threshold $\tau$ of the aggregated attention mass. This is a one-time, model-specific process, resulting in a task-agnostic configuration that generalizes across inputs.

\begin{algorithm}[t]
  \caption{\small Inference with Online Core Context Selection}
  \label{alg:dynamic_attention}
  \begin{algorithmic}[1]\small
    \REQUIRE Queries $\bQ \small{\in} \mathbb{R}^{L \times d_h}$, keys $\bK \small{\in} \mathbb{R}^{L \times d_h}$, values $\bV \small{\in} \mathbb{R}^{L \times d_h}$, block size $b$, window size $w$, configuration ${\bf p}^*$ from Alg.~\ref{alg:offline_search}.
    \STATE Compute token scores ${\bf s} \small{=} \text{softmax}({\bf Q}_{L,:} {\bf K}^\top / \sqrt{d_h})$ 
    \STATE Split $\bK$ into $m=\lfloor L/b \rfloor$ blocks (same for $\bV$)
    \STATE Compute block sparsity ${\bf h}\small{=}[h_1,\small{\dots},h_m]$ via Eq.~(\ref{eq:sparsity})
    \STATE Sort blocks in  descending order by the sparsity $\bf h$ 
    \STATE Compute token requirement ${\bf t}\small{=}[t_1, t_2,\dots,t_m]$ based on configuration ${\bf p}^*$ via Eq.~(\ref{eq:token_budget_compute})\\
    \STATE Generated selected tokens index $\mathcal{S}$ via Eq.~(\ref{eq:index_compute})
  \STATE ${\bf Att} = \text{Softmax}\left( {\bf Q} [{\bf K}^\text{G}; {\bf K}^\text{L}]^\top / {\sqrt{d_h}} \right) [{\bf V}^\text{G}; {\bf V}^\text{L}]$,~~\\$\bf K^\text{G}\small{=}{\bf K}_{\mathcal{S}}$, $\bf V^\text{G}\small{=}{\bf V}_{\mathcal{S}}$, $\bf K^\text{L}\small{=}{\bf K}_{L-w:,:}$, $\bf V^\text{L}\small{=}{\bf V}_{L-w:,:}$
    \ENSURE Attention output $\mathbf{Att}$
  \end{algorithmic}
\end{algorithm}

\subsection{Online Core Context Selection}\label{sec:online}
During inference, the offline determined configuration $\mathbf{p}^*$ for each head dictates its overall token budget. The goal of the online stage is to distribute this budget adaptively across the input sequence based on the current context, thereby reducing the context-dependent redundancy.
For an input sequence of length $L$, we first obtain a global, token-level importance score $\mathbf{s} \in \mathbb{R}^L$. We compute $\mathbf{s}$ using the query vector of the last token, which in causal self-attention has full visibility over the entire sequence and thus serves as a natural context summarizer. This allows us to identify tokens that are most relevant to the current generation. We compute the importance score by:
\begin{equation}\label{eq:token-wise_importance}
{\bs} = \text{softmax}(\frac{{\bf Q}_{L,:} {\bf K}^\top}{\sqrt{d_h}}).
\end{equation}
where $\mathbf{Q}_{L,:}$ is the query vector of the last token. Compared to methods like MInference~\cite{jiangminference} and FlexPrefill~\cite{laiflexprefill}, which use the last $k$ tokens to score importance (with $\mathcal O(kL)$ complexity), this reduces the cost to $\mathcal O(L)$ while maintaining performance. This design is not only computationally efficient but also avoids introducing handcrafted heuristics or arbitrary hyperparameters (e.g., the choice of $k$).
Then, similar to the offline phase, we partition the input sequence into non-overlapping blocks of size $b$. The tokens of the blocks that are not divisible are put into the local subset. 
For each block $\mathcal{B}_j$, we compute a redundancy score $\mathbf{h}_j$ that quantifies its information density:
\begin{equation}\label{eq:sparsity}
    \mathbf{h}_j = (1 - \alpha) \cdot \sum_{i \in \mathcal{B}_j} \mathbf{s}_i + \alpha \cdot \left(1-\frac{\sum_{i \in \mathcal{B}_j} \mathbf{s}_i^2}{\left( \sum_{i \in \mathcal{B}_j} \mathbf{s}_i \right)^2}\right),
\end{equation}
The first term (weighted by $1-\alpha$) penalizes blocks with low total attention mass. The second term (weighted by $\alpha$) is a variant of the Herfindahl-Hirschman Index~\cite{rhoades1993herfindahl}; it decreases as attention becomes more concentrated on a few tokens (indicating higher redundancy within the block). Thus, a lower $\mathbf{h}_j$ score indicates a more redundant block. The blocks are then sorted by their $\mathbf{h}_j$ scores, resulting in an ordered list of block indices $\mathbf{I} = \text{SortIndex}(\mathbf{h})$.

In online selection, we adaptively assign a token budget $\mathbf{t}_i$ to each block, dictated by its rank in $\mathbf{I}$ and the head's pre-defined configuration $\mathbf{p}^*$. Recall that $\mathbf{p}^* = [p_1, p_2, ..., p_b]$ defines the head's sparsity policy: it specifies that a proportion $p_k$ of blocks should be assigned a budget of $k$ tokens. We enforce this policy through a deterministic mapping:
\begin{equation}\label{eq:token_budget_compute}
    \mathbf{t}_i = \Psi[\mathbf{I}_i], ~\mathrm{where}~
    \Psi = \Big( \underbrace{1,\ldots, 1}_{\lfloor m \times p_1 \rfloor }, \underbrace{2, \ldots, 2}_{\lfloor m \times  p_2 \rfloor}, \ldots, \underbrace{b \ldots, b}_{\lfloor m \times  p_b \rfloor} \Big).
\end{equation}
This mechanism ensures that the most information-dense blocks (highest rank in $\mathbf{I}$) receive the largest token budgets (later entries in $\Psi$), as dictated by the head's sparsity configuration $\mathbf{p}^*$. For each block $\mathcal{B}_i$, we select the top-$\mathbf{t}_i$ tokens with the highest importance scores $\mathbf{s}$:
\begin{equation}\label{eq:index_compute}
    \mathcal{S}_i = \mathrm{Top}(\bt_i; \mathcal{B}_i).
\end{equation}
The union of all these subsets $\mathcal{S} = \bigcup_{i=1}^m \mathcal{S}_i$ forms the global subset. This set, concatenated with the local subset of the most recent $w$ tokens, is used to compute the final attention output via Eq.~(\ref{eq:atten}). As analyzed in Section~\ref{app:breakdown}, the overhead of our lightweight redundancy metric is marginal compared to the significant computational savings achieved by sparsification.

\textbf{Online Token Selection in Decoding Stage}.
Our method naturally extends to the decoding stage, where tokens are incrementally generated. 
Specifically, once the number of new tokens reaches the block size $b$, we evaluate the importance of its tokens using the same scoring function in Eq.~(\ref{eq:token-wise_importance}) based on the last query. 
We then retain only the top-$t$ most informative tokens in the block based on their scores. The value $t\small{=}\lfloor\sum_{k \in \mathcal{K}} k \cdot p_k\rfloor$ corresponds to the average number of tokens preserved per block in a head, derived from the determined configuration ${\bf p}^*$. This ensures that only the relevant context is kept, enabling efficient and effective long-sequence modeling in the decoding stage.

\subsection{Theoretical  Guarantees for \sexyname}\label{sec:theorem} 
A desirable property of any sparse attention mechanism is the ability to approximate full attention with bounded and predictable deviation. To characterize this behavior, we provide a theoretical analysis showing that the approximation error introduced by \sexyname is explicitly controlled by the retained attention mass of the selected tokens. This establishes a principled foundation for the core-context selection mechanism used in both prefilling and decoding stages.

Let $\mathbf{Att}_i$ denote the full attention output for the $i$-th query, and let $\widetilde{\mathbf{Att}}_i$ denote the corresponding approximation computed using only the selected global and local tokens. The following theorem provides a formal upper bound.
\begin{thm}[\textbf{Error Bound for a Single Query}]\label{thm:err}
Let $\gamma_i = \sum_{j \notin \mathcal{S}_{\text{total}}} \frac{\exp(\mathbf{A}_{ij})}{Z_i}$ be the total probability mass that the full softmax assigns to tokens omitted by \sexyname. Here, $\mathcal{S}_{\text{total}}$ denotes the complete set of token indices selected by our method, including the global subset and local subset. Then the approximation error satisfies:
\begin{equation*}
| \mathbf{Att}_i - \widetilde{\mathbf{Att}}_i |1 \leq 2\gamma_i \cdot |\mathbf{V}|_\infty,
\end{equation*}
where $|\mathbf{V}|_\infty$ is the maximum absolute entry in value matrix.
\end{thm}
Theorem \ref{thm:err} shows that the error scales linearly with $\gamma_i$, \ie, the fraction of probability mass on unselected tokens. Since the offline sparsity Determination in \sexyname enforces that the aggregated retained attention mass exceeds a threshold $\tau$, we effectively ensure $\gamma_i \leq 1 - \tau$. Thus, the bound is directly governed by a tunable hyperparameter, providing explicit controllability. Furthermore, the guaranteed inclusion of the most recent local tokens prevents large $\gamma_i$ values in generative decoding, while head-specific sparsity budgets ensure that redundancy is eliminated only where error tolerance is naturally higher. 

The full proof of Theorem~\ref{thm:err}, along with additional derivations and discussion, is provided in Appendix A. Together, these results establish that \sexyname is a principled approximation mechanism whose deviation from full attention is not only bounded, but also systematically regulated by its sparsity configuration process.

\begin{table*}[t]
\centering
\caption{Comparison on LongBench-E~\cite{bai2023longbench} across efficient-attention methods under LLaMA3.1-8B and Qwen2.5-7B models. We report the attention computation latency at 64K contexts.}
\label{tab:longbench}
\renewcommand{\tabcolsep}{5.7pt}
\begin{tabular}{l|cccccc|c|c}
\hline
Methods & Sin. Doc. QA & Mul. Doc. QA & Sum. & Few Shot & Syn. & Code & ~Average~ & Latency (ms) \\ \hline

LLaMA3.1-8B-Instruct & 51.63 & 53.23 & 30.78 & 68.67 & 54.29 & 60.52 & 53.19 & 316.14 \\
~~$\bullet~$MInference~\cite{jiangminference} & 51.70 & 52.72 & 30.76 & 68.58 & 53.50 & 61.12 & 53.06 & 324.84  \\
~~$\bullet~$FlexPrefill~\cite{laiflexprefill} & 50.35 & 52.85 & 30.71 & 68.41 & 54.30 & 62.76 & 53.23 & 280.11 \\
~~$\bullet~$XAttention~\cite{xu2025xattention} & 49.96 & 51.98 & 31.22 & 68.07 & 48.50 & 55.75 & 50.91 & 133.85 \\
~~$\bullet~$\sexynamellm (Ours) & 52.28 & 52.83 & 30.84 & 68.40 & 54.86 & 61.82 & \textbf{53.51} & \textbf{120.96}\\
\hline

Qwen2.5-7B-Instruct & 48.75 & 52.24 & 27.81 & 65.00 & 52.00 & 61.14 & 51.16 & 268.55\\
~~$\bullet~$MInference~\cite{jiangminference} & 48.80 & 52.37 & 27.64 & 64.67 & 51.50 & 62.08 & 51.18 & 292.33 \\
~~$\bullet~$FlexPrefill~\cite{laiflexprefill} & 49.08 & 52.16 & 27.86 & 65.18 & 52.00 & 62.20 & 51.41 & 244.37\\
~~$\bullet~$XAttention~\cite{xu2025xattention} & 48.50 & 50.08 & 27.48 & 66.40 & 50.50 & 60.98 & 50.66 & 119.87 \\
~~$\bullet~$\sexynamellm (Ours) & 48.50 & 52.91 & 27.74 & 64.92 & 52.25 & 62.74 & \textbf{51.51} & 105.93\\ 
\hline
\end{tabular}
\end{table*}

\begin{table*}[t]
\centering
\caption{Comparison on RULER~\cite{hsieh2024ruler} across efficient-attention methods under LLaMA3.1-8B and Qwen2.5-7B models. We report the attention computation latency at 128K contexts.}\label{tab:ruler}
\renewcommand{\tabcolsep}{9.7pt}
\begin{tabular}{l|cccccc|c|cc}
\hline
Methods & 4K & 8K & 16K & 32K & 64K & 128K & ~Average~ & Latency (ms) \\ \hline

LLaMA3.1-8B-Instruct & 96.74 & 94.03 & 92.02 & 84.17 & 81.32 & 76.89 & 87.52 & 1282.05  \\
~~$\bullet~$MInference~\cite{jiangminference} & 96.54 & 94.06 & 91.37 & 85.79 & 83.03 & 54.13 & 84.15 & 837.09  \\
~~$\bullet~$FlexPrefill~\cite{laiflexprefill} & 95.99 & 93.67 & 92.73 & 88.14 & 81.14 & 74.67 & 87.72 & 1021.47  \\
~~$\bullet~$XAttention~\cite{xu2025xattention} & 96.15 & 93.95 & 93.71 & 90.90 & 83.35 & 72.57 & 88.44 & 504.00 \\
~~$\bullet~$\sexynamellm (Ours) & 96.31 & 95.38 & 93.92 & 86.38 & 82.89 & 77.46 & \textbf{88.72} & 453.22  \\
\hline

Qwen2.5-7B-Instruct & 96.00 & 94.85 & 91.77 & 89.85 & 70.38 & 52.92 & 82.63 & 1101.01 \\
~~$\bullet~$MInference~\cite{jiangminference} & 96.08 & 94.92 & 91.69 & 89.92 & 70.46 & 51.62 & 82.45 &  771.61 \\
~~$\bullet~$FlexPrefill~\cite{laiflexprefill} & 95.62 & 94.31 & 92.00 & 88.23 & 70.23 & 52.15 & 82.09 & 892.55  \\
~~$\bullet~$XAttention~\cite{xu2025xattention} & 95.45 & 92.91 & 92.04 & 88.84 & 68.84 & 55.36 & 82.24 & 441.36  \\
~~$\bullet~$\sexynamellm (Ours) & 96.00 & 94.77 & 91.92 & 90.08 & 69.85 & 52.31 & \textbf{82.49} & 396.72  \\
\hline

\end{tabular}%
\end{table*}

\begin{table}[t]
\caption{Characteristic comparisons between existing methods and our \sexyname.}
\newcommand{\tabincell}[2]{\begin{tabular}{@{}#1@{}}#2\end{tabular}}
\label{tab:characteristic}
\begin{center}
\begin{threeparttable}
\renewcommand{\tabcolsep}{4.5pt}
\begin{tabular}{l|ccccc}
\hline
Methods & Training-free & Dynamic & Prefilling & Decoding \\ \hline
MInference~\cite{jiangminference} & \checkmark & \checkmark & \checkmark &  \\
FlexPrefill~\cite{laiflexprefill} & \checkmark & \checkmark & \checkmark &  \\
XAttention~\cite{xiaoduoattention} & \checkmark & \checkmark & \checkmark &  \\
SnapKV~\cite{li2024snapkv} & \checkmark & \checkmark & & \checkmark  \\
CAKE~\cite{qincake} & \checkmark & \checkmark & & \checkmark  \\ 
DuoAttention~\cite{xiaoduoattention} & & & \checkmark & \checkmark  \\
\hline
\sexyname (Ours) & \checkmark & \checkmark & \checkmark & \checkmark \\ \hline
\end{tabular}
\end{threeparttable}
\end{center}
\end{table}

\section{Experiments}\label{sec:exp}
\subsection{Experimental Setup}

\textbf{Models}. We evaluate our method on two state-of-the-art large language models: \textbf{Qwen2.5-7B-Instruct}~\cite{yang2024qwen2_5} and \textbf{LLaMA-3.1-8B-Instruct}~\cite{meta2024llama3}. Both support context windows of up to 128K tokens and employ grouped-query attention (GQA). Qwen2.5-7B-Instruct demonstrates enhanced capabilities in multilingual understanding, mathematical reasoning, and code generation. LLaMA-3.1-8B-Instruct is optimized for dialogue and generation tasks across multiple languages.

\textbf{Compared Methods}. As a training-free sparse attention method, \sexyname is primarily compared with three state-of-the-art training-free attention methods: \textbf{1) Minference}~\cite{jiangminference}: This method employs offline determination to select optimal sparse attention patterns per attention head, combined with online dynamic adjustment of computation regions for each pattern. We use the officially released implementation in all experiments. \textbf{2) FlexPrefill}~\cite{laiflexprefill}: This approach dynamically selects between Query-Aware and Vertical-Slash attention patterns per head, while adaptively determining the required Key-Value indices for computation. In our experiments, we use the official implementation and follow the original paper's settings: $\gamma$=0.9, $\tau$=0.1, with a minimum retained budget of 1,024 tokens and a block size of 128 across all evaluated models. \textbf{3) XAttention}~\cite{xu2025xattention}: This method employs an antidiagonal scoring pattern to select sparse attention blocks, reducing computation by processing only the selected regions. In our experiments, we follow the original paper's optimal setting with stride $S=8$. For LLaMA3.1-8B, we adopt the officially released set of minimum thresholds to determine block selection. For Qwen2.5-7B, we set the threshold to 0.9 to preserve more contextual information. We summarize differences from existing methods in Table~\ref{tab:characteristic}. This constitutes a fair comparison within the same category of techniques.

\textbf{Implementation Details}. We integrate the proposed \sexyname with existing LLMs through a \textbf{plug-and-play} replacement of full self-attention, requiring \textbf{no architectural modifications} or \textbf{parameter updates}. Experiments run on a server with 8$\times$ NVIDIA A800 GPUs (80GB VRAM each). All the latency is measured on a single A800 GPU. All latency for full attention are based on the highly optimized FlashAttention2~\cite{dao2024flashattention} kernel. To align with established practices for long-context evaluation~\cite{jiangminference,laiflexprefill,xu2025xattention}, latency is measured at batch size = 1 with input sequences up to 128K tokens. The offline sparsity threshold is set to $\tau = 0.9$. Block size $b$ and window size $w$ are set to 128 and 4096 for all models, respectively. For sparsity configuration determination, we adopt the 14 candidate configurations for all LLMs. These candidate configurations are generated using Gaussian distributions (Eq.~\ref{eq:gaussian}) with $\mu$ sampled from $\log_2(1)$ to $\log_2(128)$. The optimal configuration per head is determined via the offline procedure in Section~\ref{sec:offline}. All hyperparameters are held constant across experiments to ensure reproducibility. 

\textbf{Datasets}. We conduct a comprehensive evaluation across long-context modeling, short-context understanding, complex reasoning, and multi-turn dialogue tasks. For long-context assessment, we employ \textbf{LongBench-E}~\cite{bai2023longbench}, which includes 14 diverse tasks for long-context understanding, and \textbf{RULER}~\cite{hsieh2024ruler}, a synthetic benchmark built on the Needle-in-a-Haystack paradigm. For short-context evaluation, we use \textbf{MMLU}~\cite{hendrycks2021measuring} (multi-subject knowledge), \textbf{GSM8K}~\cite{cobbe2021gsm8k} (mathematical reasoning), and \textbf{HumanEval}~\cite{chen2021evaluating} (code generation). Additionally, we evaluate on \textbf{OlympiadBench}~\cite{he2024olympiadbench} for advanced scientific reasoning using Olympiad-level problems. We also evaluate on \textbf{MT-Bench-01}~\cite{bai2024mt} to validate the effectiveness of \sexyname in multi-turn conversations. Refer to Appendix C for more details.

\begin{table*}[t]
\begin{minipage}{0.48\textwidth}
    \centering
    \caption{Comparisons on short-context tasks.}\label{tab:short_tasks}
    \begin{tabular}{l|ccc}
\hline
Methods & MMLU & GSM-8K & HumanEval \\ \hline

LLaMA3.1-8B-Instruct & \textbf{69.38} & 83.85 & \textbf{68.29}  \\
~~$\bullet~$MInference~\cite{jiangminference} & 69.14 & 84.08 & 67.30  \\
~~$\bullet~$FlexPrefill~\cite{laiflexprefill} & 69.16 & \underline{84.15} & 67.07 \\
~~$\bullet~$XAttention~\cite{xu2025xattention} & \underline{69.21} & \underline{84.15} & 67.39 \\
~~$\bullet~$\sexynamellm (Ours) & \underline{69.21} & \textbf{84.23} & \underline{67.46} \\ \hline
Qwen2.5-7B-Instruct & 74.22 & 79.68 & \textbf{81.71}  \\
~~$\bullet~$MInference~\cite{jiangminference} & 74.14 & 80.29 & 79.88  \\
~~$\bullet~$FlexPrefill~\cite{laiflexprefill} & \underline{74.23} & \underline{80.36} & \underline{81.10} \\
~~$\bullet~$XAttention~\cite{xu2025xattention} & 74.20 & 79.30 & 80.49 \\
~~$\bullet~$\sexynamellm (Ours) & \textbf{74.26} & \textbf{80.44} & \underline{81.10} \\ \hline

\end{tabular}
\end{minipage}
\hfill
\begin{minipage}{0.48\textwidth}
\centering
\caption{Comparisons on OlympiadBench.}\label{tab:olympiad}
\begin{tabular}{l|cc|c}
\hline
Methods & Math & Physics & Avg. \\ \hline
LLaMA3.1-8B-Instruct & 13.84 & 8.85 & 11.84 \\
~~$\bullet~$MInference~\cite{jiangminference} & 13.64 & 9.73 & 12.08 \\
~~$\bullet~$FlexPrefill~\cite{laiflexprefill} & 13.39 & 8.85 & 11.57 \\
~~$\bullet~$XAttention~\cite{xu2025xattention} & \textbf{13.71} & 9.30 & 11.95 \\
~~$\bullet~$\sexyname (Ours)~~~~~~~ & ~~13.19~~ & ~~\textbf{10.59}~~ & ~~\textbf{12.15}~~ \\ \hline
Qwen2.5-7B-Instruct & 38.85 & 19.73 & 31.20 \\
~~$\bullet~$MInference~\cite{jiangminference} & 38.79 & 19.28 & 30.99 \\
~~$\bullet~$FlexPrefill~\cite{laiflexprefill} & 38.73 & 19.06 & 30.86 \\
~~$\bullet~$XAttention~\cite{xu2025xattention} & \textbf{38.94} & 18.84 & 30.90 \\
~~$\bullet~$\sexyname (Ours) & 38.82 & \textbf{19.95} & \textbf{31.27} \\ \hline
\end{tabular}
\end{minipage}
\end{table*}

\begin{table*}[t]
    \centering
\caption{Results on MT-Bench-101 with Qwen2.5-7B-Instruct.}\label{tab:mtbench}
\renewcommand{\tabcolsep}{5.3pt}
\begin{tabular}{l|ccccccccccccc|c} 
\hline
\multicolumn{1}{c|}{Methods} & GR & IC & AR & FR & MR & CC & TS & CR & SA & SI & CM & PI & SC & Average \\ 
\hline
Qwen2.5-7B-Instruct & 8.20 & \textbf{7.71} & 9.49 & 9.56 & \textbf{7.57} & 9.89 & 8.90 & 9.50 & \textbf{9.29} & 8.26 & 9.19 & \textbf{8.64} & 9.42 & 8.90 \\
~~$\bullet$~FlexPrefill~\cite{laiflexprefill} & 7.81 & 7.67 & 9.55 & 9.57 & 7.23 & 9.91 & \textbf{9.40} & \textbf{9.57} & 9.22 & 8.33 & \textbf{9.27} & 8.63 & \textbf{9.53} & 8.90 \\
~~$\bullet$~\sexyname (Ours) & \textbf{8.50} & \textbf{7.71} & \textbf{9.56} & \textbf{9.74} & 7.48 & \textbf{9.91} & 9.39 & 9.49 & 9.26 & \textbf{8.38} & 9.20 & 8.62 & 9.43 & \textbf{8.97} \\
\hline
\end{tabular}
\end{table*}

\begin{table*}[t]
\centering
\caption{Performance comparisons across prefilling and decoding on LongBench-E.}\label{tab:end_to_end}
\renewcommand{\tabcolsep}{8.5pt}
\begin{tabular}{l|cccccc|c}
\hline
Methods & Sin. Doc. QA & Mul. Doc. QA & Sum. & Few Shot & Syn. & Code & ~Average~ \\ \hline

MInference~\cite{jiangminference} + SnapKV~\cite{li2024snapkv} & 49.66 & 53.16 & 29.90 & 67.60 & 51.72 & 58.25 & 51.72
\\
FlexPrefill~\cite{laiflexprefill} + CAKE~\cite{qincake} & 50.27 & 52.87 & 30.77 & 66.96 & 52.25 & 60.24 & 52.23
\\
DuoAttention~\cite{xiaoduoattention} & 51.05 & 52.70 & 29.70 & 67.27 & 52.25 & 59.86 & 52.14 \\
\sexynamellm (Ours) & 50.71 & 53.11 & 30.24 & 67.93 & 52.59 & 60.05 & \textbf{52.44}
\\
\hline
\end{tabular}
\end{table*}

\subsection{Comparisons with State-of-the-art Methods}
\textbf{Comparisons on LongBench-E.}
In Table~\ref{tab:longbench}, \sexyname achieves the highest average score among sparse attention methods. For LLaMA3.1-8B, it attains a \textbf{2.6$\times$} speedup (120.96ms vs. 316.14ms) at 64K context while outperforming XAttention by 2.6 average score. On Qwen2.5-7B, it delivers the best performance with the fastest latency (2.5$\times$ speedup). These results validate our method's effectiveness and architectural adaptability without compromising accuracy.

\textbf{Comparisons on RULER.}
In Table~\ref{tab:ruler}, \sexyname achieves superior efficiency and accuracy in long context understanding across different context lengths. On LLaMA3.1-8B-Instruct and Qwen2.5-7B-Instruct, \sexyname attains the highest average score with up to \textbf{2.8$\times$} speedup compared to vanilla self-attention (0.45s vs 1.28s), outperforming the strongest counterpart method XAttention despite faster computation. These results confirm that our dynamic sparsity mechanism effectively maintains performance while accelerating computation across varying sequence lengths.

\textbf{Comparisons on Short-context Tasks}.
In Table~\ref{tab:short_tasks}, our \sexyname achieves competitive results on LLaMA3.1-8B-Instruct-128K and outperforms all baselines on Qwen2.5-7B-Instruct-128K, demonstrating robust cross-model generalization. Notably, it preserves the capability of the original model, validating that our method effectively retains critical information without fine-tuning.

\textbf{Comparisons on Reasoning Tasks}.
To evaluate the effectiveness of our \sexyname on challenging reasoning tasks requiring long-context understanding, we conduct experiments on the OlympiadBench benchmark, which includes advanced problems in Mathematics and Physics. As shown in Table~\ref{tab:olympiad}, our method demonstrates competitive and often superior performance compared to existing sparse attention methods. On both of LLaMA3.1-8B-Instruct and Qwen2.5-7B-Instruct, \sexyname achieves the highest average score, outperforming all compared methods. Notably, \sexyname obtains the highest score in Physics, while maintaining competitive performance in Mathematics. This result further confirms the strong capability of \sexyname in handling complex scientific reasoning. 

\textbf{Comparisons on Multi-turn Conversation}. To evaluate the model's capability in maintaining contextual coherence and reasoning across multi-turn interactions, we conduct experiments on MT-Bench-101~\cite{bai2024mt}. As shown in Table~\ref{tab:mtbench}, \sexyname achieves an average score of 8.97, surpassing both the full-attention baseline (8.90) and strong sparse methods such as FlexPrefill (8.90). This advantage is consistent across key conversational dimensions, including generation, reasoning, and safety. The results demonstrate that our \sexyname effectively preserves essential dialogue context across turns, validating its robustness and practicality for real-world conversational applications.

\begin{figure*}[t]
    \centering
    \includegraphics[width=0.96\linewidth]{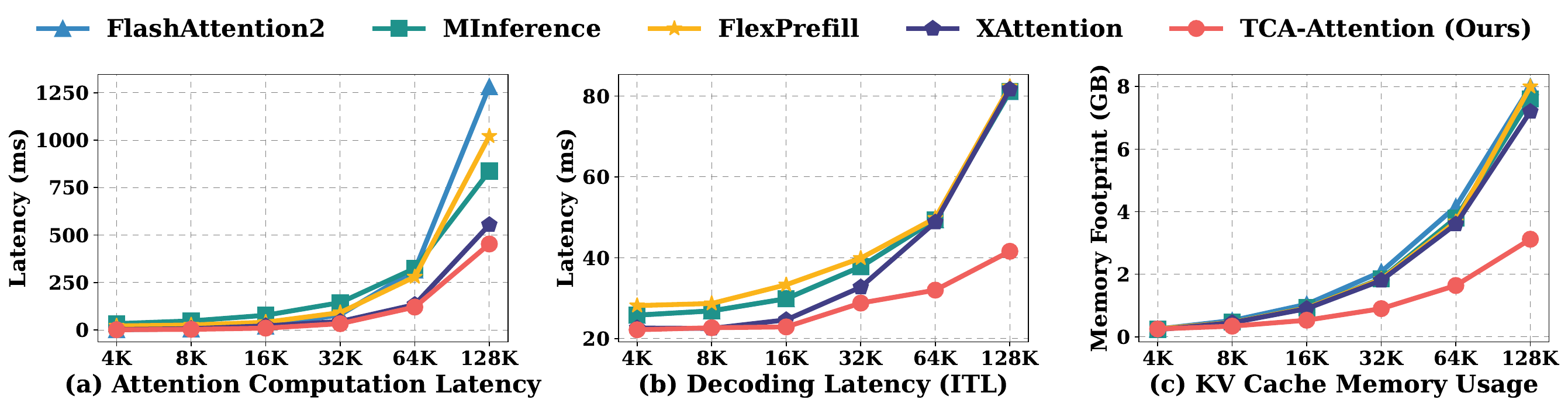}
    \caption{Comparisons in terms of computational and storage overhead on LLaMA3.1-8B-Instruct.
  Attention computation latency is the time to compute a single attention layer. ``ITL'' (inter token latency) is the time between generating consecutive tokens (except for the first token) in decoding.}\label{fig:efficiency}
\end{figure*}

\textbf{Comparisons on Computational Efficiency}.
As shown in Figure~\ref{fig:efficiency}, \sexyname achieves a \textbf{2.8$\times$} speedup in prefilling latency and a \textbf{2.1$\times$} faster decoding speed over vanilla self-attention at 128K context length on LLaMA3.1-8B-Instruct. By dynamically selecting tokens, it also reduces KV cache memory by \textbf{61\%} (3.12GB vs. 8.00GB). In contrast, methods like MInference only accelerate prefilling, leaving decoding latency and memory unchanged. This demonstrates \sexyname's unique advantage in enabling efficiency in LLMs inference. These results confirm that \sexyname excels in both prefill and decoding stages while substantially reducing memory overhead, a crucial advantage for deploying LLMs in long-context scenarios.

\begin{figure*}[t]
\centering
\begin{subfigure}[h]{0.28\linewidth}
    \includegraphics[width=\textwidth]{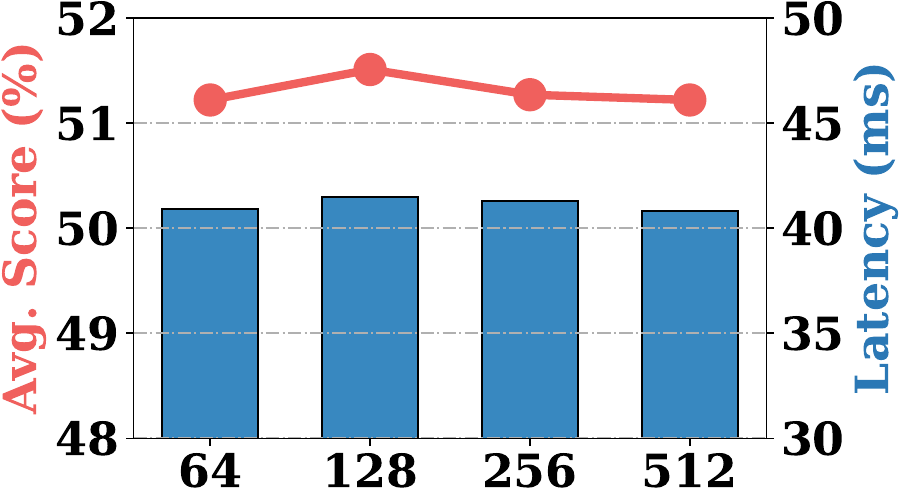}
    \caption{Ablation on block size $b$.}
    \label{fig:abl_block}
\end{subfigure}
\hfill
\begin{subfigure}[h]{0.29\linewidth}
    \includegraphics[width=\textwidth]{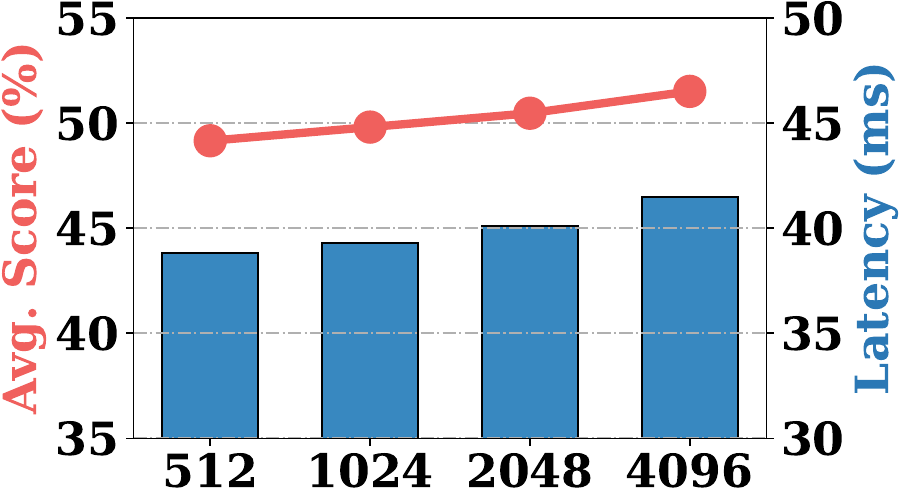}
    \caption{Ablation on window size $w$.}
    \label{fig:abl_window}    
\end{subfigure}
\hfill
\begin{subfigure}[h]{0.33\textwidth}
    \includegraphics[width=\textwidth]{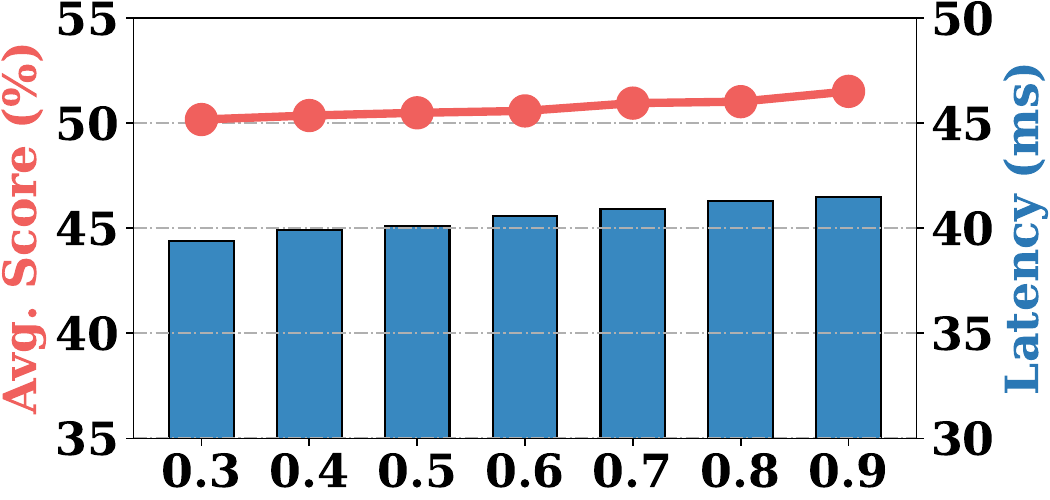}
    \caption{Ablation on thresholds $\tau$.}
    \label{fig:abl_tau}
\end{subfigure}
\caption{Ablations on hyperparameters. We report the average score on LongBench-E and the decoding latency.}\label{fig:ablations}
\end{figure*}

\textbf{Performance Comparison across Prefilling and Decoding}.  
To validate the effectiveness of our unified framework across \textit{both prefilling and decoding stages}, we further conduct experiments on LongBench-E by combining the strongest prefilling-stage baselines with KV cache compression methods. The results in Table~\ref{tab:end_to_end} demonstrate that our approach achieves superior performance across most tasks. Although DuoAttention aims to reduce latency in both prefilling and decoding stages, the official DuoAttention implementation employs a standard FlashAttention-2 kernel for prefilling, thus failing to reduce computation in this computationally intensive stage. This highlights that our method establishes a more efficient acceleration method for long-context processing.

\subsection{Ablation Studies}\label{sec:ablation}

We perform ablation studies on Qwen2.5-7B-Instruct-128K and report the average score on LongBench-E. Refer to Appendix D for more ablations.

\textbf{Ablations on Block Size $b$}. As shown in Figure~\ref{fig:abl_block}, the model achieves the highest average score of 51.51 with a block size of 128. Smaller block sizes may fail to capture sufficient contextual redundancy, while larger sizes risk over-compressing critical information.

\textbf{Ablations on Local Window Size $w$}. 
As shown in Figure~\ref{fig:abl_window}, the performance of the model improves consistently as the window size increases, achieving the highest average score of 51.51 with a window size of 4096. Expanding the window size yields consistent performance gains by preserving richer local context details, with latency increasing only marginally from 38.8ms to 41.5ms.

\begin{figure*}[t]
    \centering
    \includegraphics[width=1.0\linewidth]{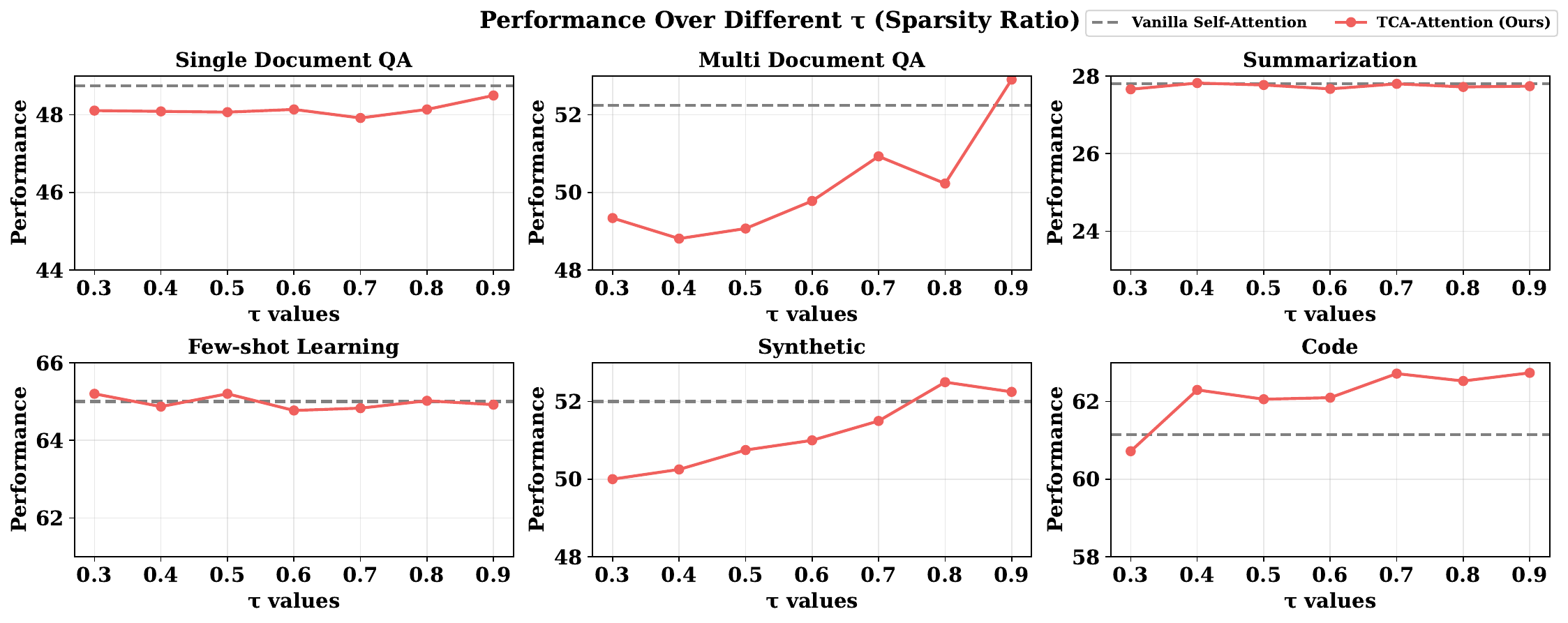}
    \caption{Performance comparison across different $\tau$ values (corresponding to different sparsity ratios) on LongBench-E tasks.}
    \label{fig:task_tau}
\end{figure*}

\textbf{Ablations on Threshold $\tau$}. 
We analyze the impact of the threshold $\tau$ on model performance, where lower $\tau$ values correspond to higher compression rates. As shown in Figure~\ref{fig:abl_tau}, performance generally improves as $\tau$ increases from 0.3 to 0.9, with peak accuracy attained at $\tau=0.9$. This reflects that retaining more contextual information benefits overall effectiveness, while latency increases only marginally. 
Detailed per-task results (Figure~\ref{fig:task_tau}) further reveal that tasks relying on localized context, such as Single-Document QA and Summarization, remain robust even under high compression ($\tau=0.3$, 84.69\% compression), suffering only a slight average drop. In contrast, tasks requiring global reasoning across the sequence, such as Multi-Document QA and Code, are more sensitive to aggressive compression. These observations suggest a practical guideline: conservative thresholds ($\tau=0.6$ or $0.9$) are advisable for globally-dependent tasks, whereas locally-dependent tasks can tolerate stronger compression to maximize efficiency with minimal performance loss.

\textbf{Ablation Study on Balancing Parameter $\alpha$}. We analyze the sensitivity of \sexyname to the balance parameter $\alpha$ in our redundancy metric. Results in Table~\ref{tab:abl_alpha} show consistent performance across a wide $\alpha$ range (0.1–0.9), demonstrating strong robustness. The optimal average score (51.51 on LongBench-E) occurs at $\alpha=0.5$, while even extreme values ($\alpha=0.1$ or $\alpha=0.9$) cause minimal degradation ($\leq 0.38$). This insensitivity to precise tuning confirms the practical deployability of our method.

\textbf{Ablation Study on Calibration Dataset}. \revise{We conduct an ablation study to evaluate \sexyname's sensitivity to calibration dataset choice. As shown in Table~\ref{tab:calibration}, our method achieves consistent performance across different calibration datasets and sizes. When calibrated on diverse domains, including general web text (SlimPajama), long-form governmental documents (GovReport), and programming code (McEval). The average score on LongBench-E remains remarkably stable (51.54-51.56). More importantly, this stability persists even with minimal calibration: using just a single input sequence (size=1), our method achieves virtually identical performance when randomly sampling five samples, with variations of $\leq 0.04$. These results demonstrate two critical properties: 1) head-specific sparsity transfers well across domains, and 2) the determined configurations are remarkably stable across different inputs within the same model, requiring only minimal calibration data. This makes \sexyname highly practical for real-world deployment, as it eliminates the need for domain-specific or extensive calibration datasets while maintaining promising performance.}

\begin{table}[t]
\centering
\caption{Performance with different $\alpha$ values.}\label{tab:abl_alpha}
\begin{tabular}{c|ccccc}
\hline
alpha & 0.1 & 0.3 & 0.5 & 0.7 & 0.9 \\ \hline
Avg. & 51.14 & 51.25 & 51.51 & 51.24 & 51.13 \\ \hline
\end{tabular}
\end{table}

\begin{table}[t]
\centering
\caption{Performance with different calibration datasets.}\label{tab:calibration}
\begin{tabular}{cc|c}
\hline
Datasets & Domain & Average Score \\ \hline
SlimPajama & General web text & 51.54$_{\pm 0.03}$ \\
GovReport & Governmental docs & 51.56$_{\pm 0.04}$ \\
McEval & Programming code & 51.55$_{\pm 0.02}$ \\ \hline
\end{tabular}
\end{table}

\subsection{Latency Breakdown Analysis}\label{app:breakdown}
\begin{figure}[tb]
    \centering
    \includegraphics[width=\linewidth]{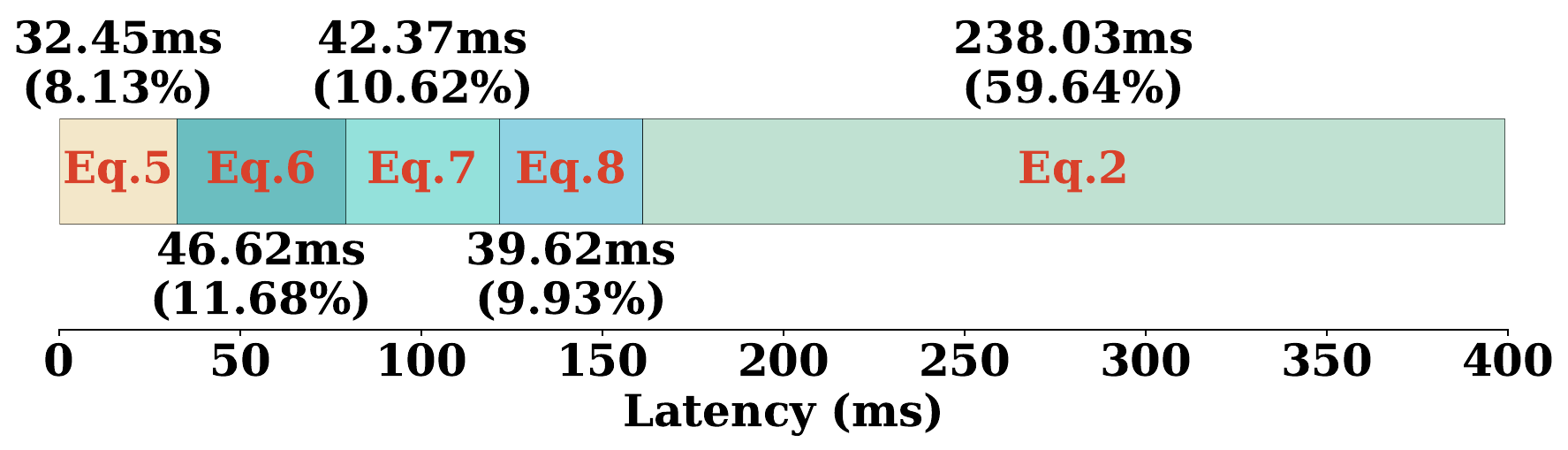}
    \caption{Latency breakdown of \sexyname components at 128K context length with a total 399.09ms.}
    \label{fig:breakdown}
\end{figure}

To validate the practical efficiency of \sexyname, we provide a detailed latency breakdown of our Triton implementation at 128K context length, as shown in Figure~\ref{fig:breakdown}. The importance score calculation (Eq.~\ref{eq:token-wise_importance}, 8.13\%) and other auxiliary operations, including redundancy metric computation (Eq.~\ref{eq:sparsity}), sorting (Eq.~\ref{eq:token_budget_compute}), and token selection (Eq.~\ref{eq:index_compute}, ), collectively contribute less than 40\% of total latency. The latency is dominated by attention computation over fused global-local subsets (Eq.~\ref{eq:atten}). Crucially, this marginal overhead is significantly outweighed by the computational savings achieved through sparsification, ultimately enabling an speedup of 2.8$\times$ compared to full attention. The efficiency primarily stems from our block-level design, which minimizes the cost of dynamic token selection and redundancy estimation, as opposed to more expensive token-level processing.

\section{Conclusion}
In this paper, we propose \methodname (\sexyname), a sparse attention mechanism that enables efficient long-context modeling without any training or architectural changes. It employs two lightweight phases, \ie, an offline head-specific sparsity determination phase and an online core context selection phase, to achieve input-adaptive sparsity.
Theoretical analysis guarantees bounded approximation error, and extensive experiments validate that \sexyname achieves significant speedup (up to 2.8$\times$) and KV cache reduction (61\%) at 128K context while maintaining competitive performance across diverse benchmarks. The plug-and-play nature of \sexyname makes it a practical solution for LLMs in long-context scenarios without retraining or architectural changes.

\clearpage

\bibliography{tpami}
\bibliographystyle{IEEEtran}
\vspace{-35pt}

\begin{IEEEbiography}[{\includegraphics[width=1in,height=1.25in,clip,keepaspectratio]{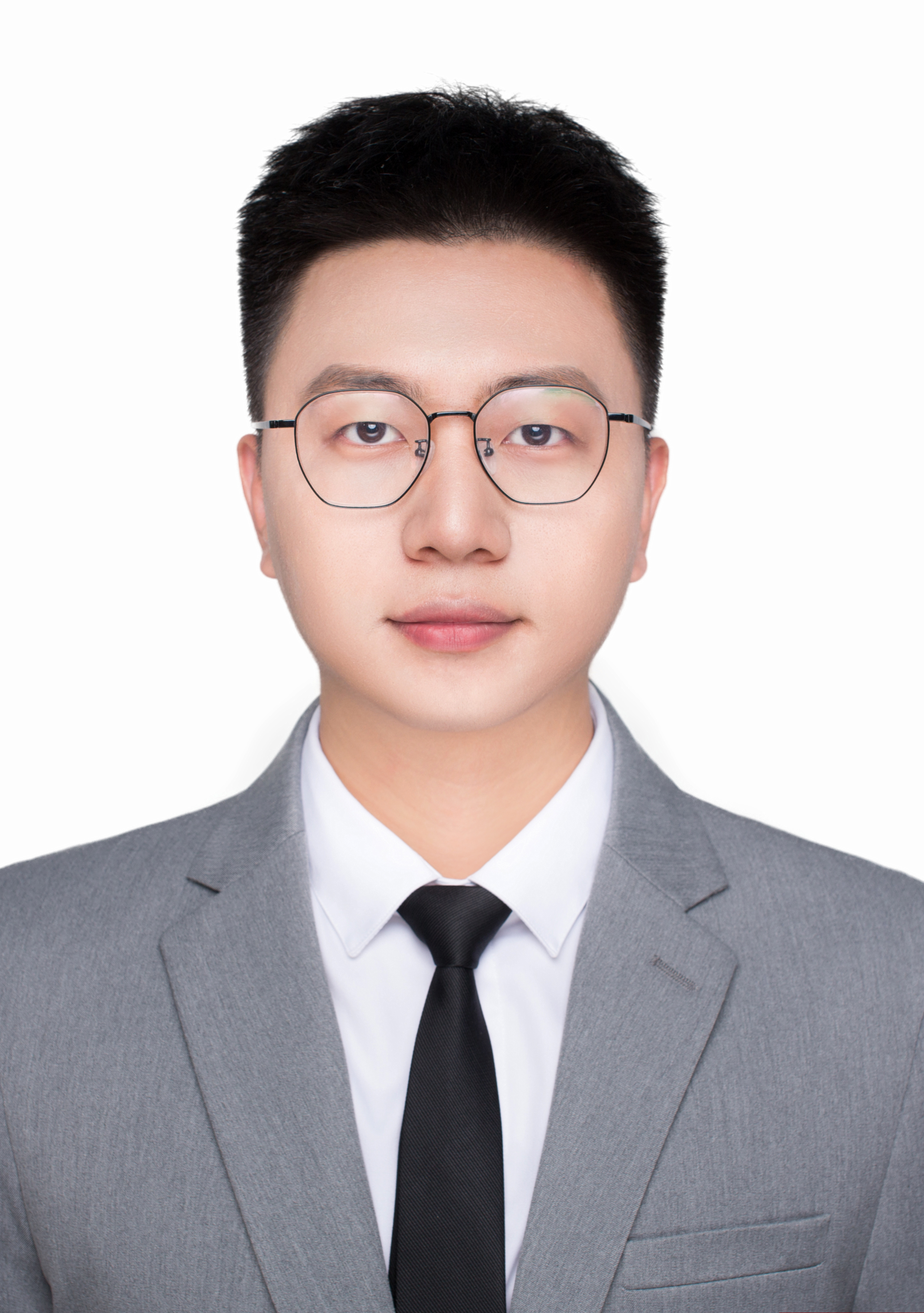}}]{Zeng You}
received the B.E. degree in Software Engineering from the School of Software Engineering, South China University of Technology, China, in 2020. He is currently working toward the Ph.D. degree in the School of Future Technology, South China University of Technology, China. His research interests include deep learning, video understanding, and efficient attention. He has published papers in top venues, including ICML, AAAI, IEEE TIP, and IEEE TCSVT. He has been invited as a reviewer for NeurIPS, AAAI, TII, and TCSVT.
\end{IEEEbiography}

\begin{IEEEbiography}[{\includegraphics[width=1in,height=1.25in,clip,keepaspectratio]{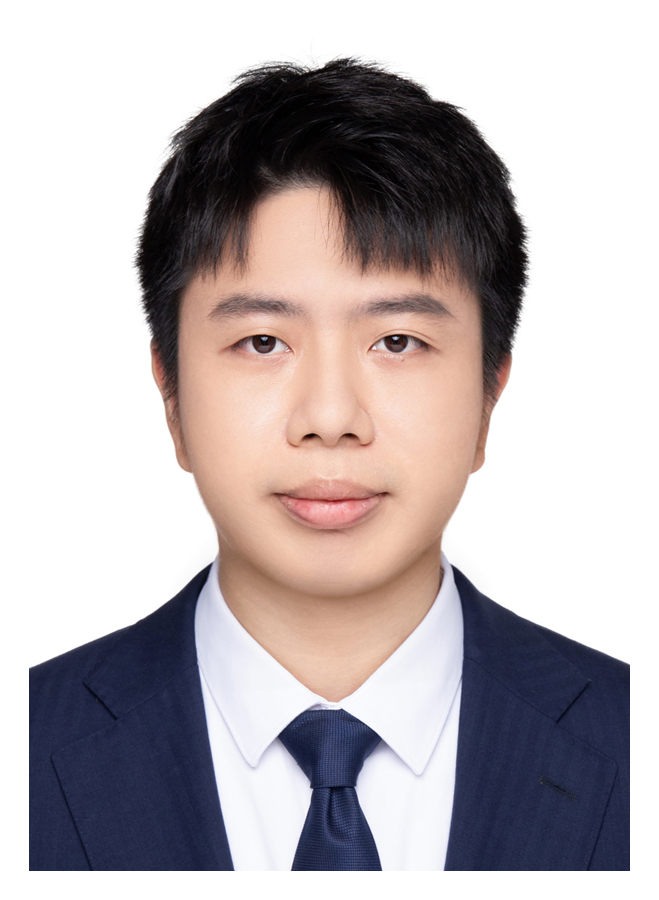}}]{Yaofo Chen}
is Post-doctoral Researcher with School of Future Technology  at South China University of Technology. He received his Ph.D. degree in the School of Software Engineering in 2024 from South China University of Technology in Guangzhou, China. His research interests include neural architecture search and test-time adaptation. He has published papers in top venues, including ICML, ICLR, CVPR, AAAI, IEEE TCSVT and Neural Networks. He has been invited as a reviewer for top-tier conferences including ICLR, ICML, NeurIPS, CVPR, ICCV, ECCV and AAAI.
\end{IEEEbiography}

\begin{IEEEbiography}[{\includegraphics[width=1in,height=1.25in,clip,keepaspectratio]{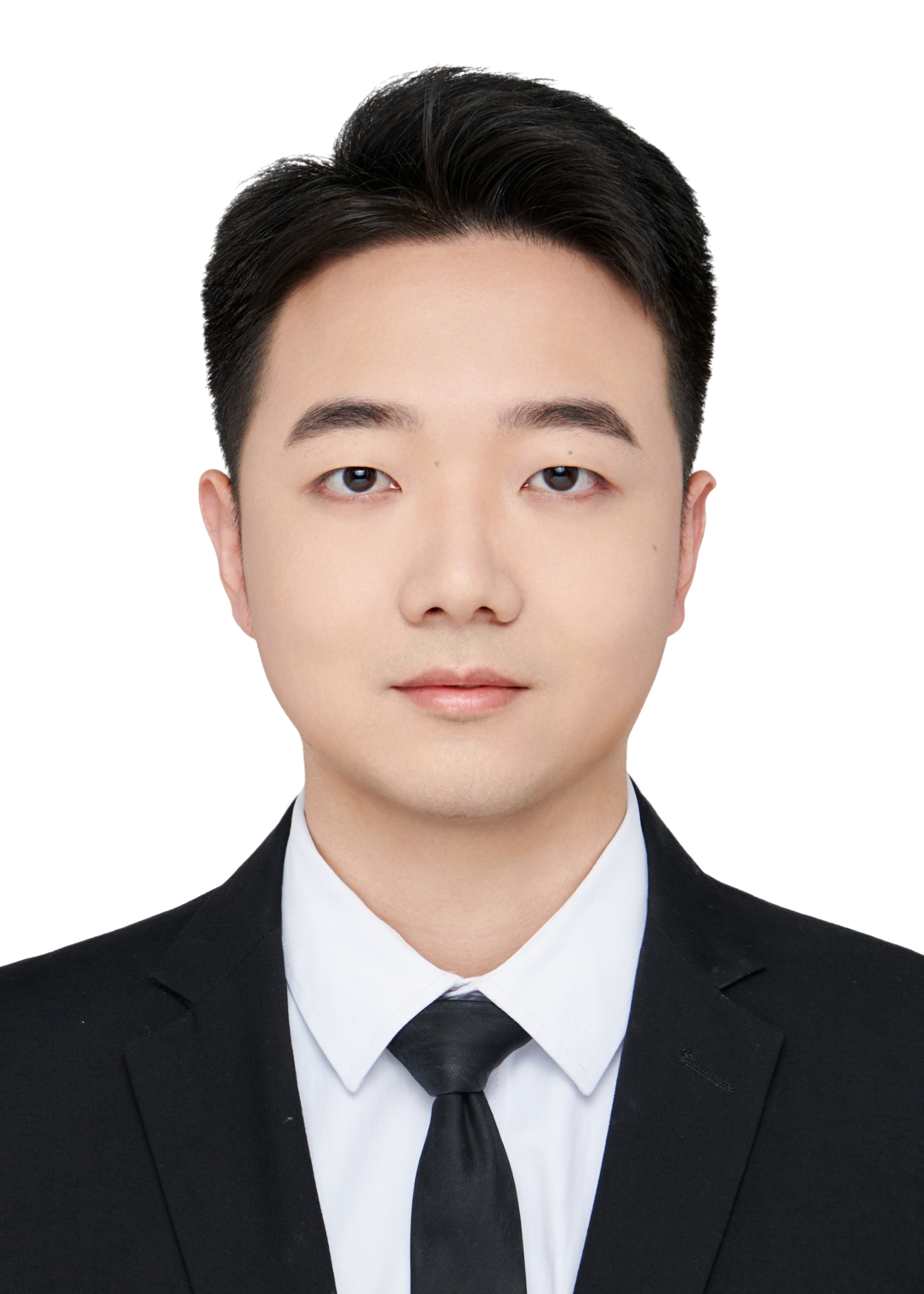}}]{Shuhai Zhang}
is currently a Ph.D. candidate at South China University of Technology, China. His research interests are broadly in machine learning and mainly focus on large language model, model compression, and adversarial robustness. He has published papers in IEEE TIP, Neural Networks, T-CSVT, NeurIPS, ICCV, ICML, ICLR, CVPR.
\end{IEEEbiography}

\vspace{-25pt}
\begin{IEEEbiography}[{\includegraphics[width=1in,height=1.25in,clip,keepaspectratio]{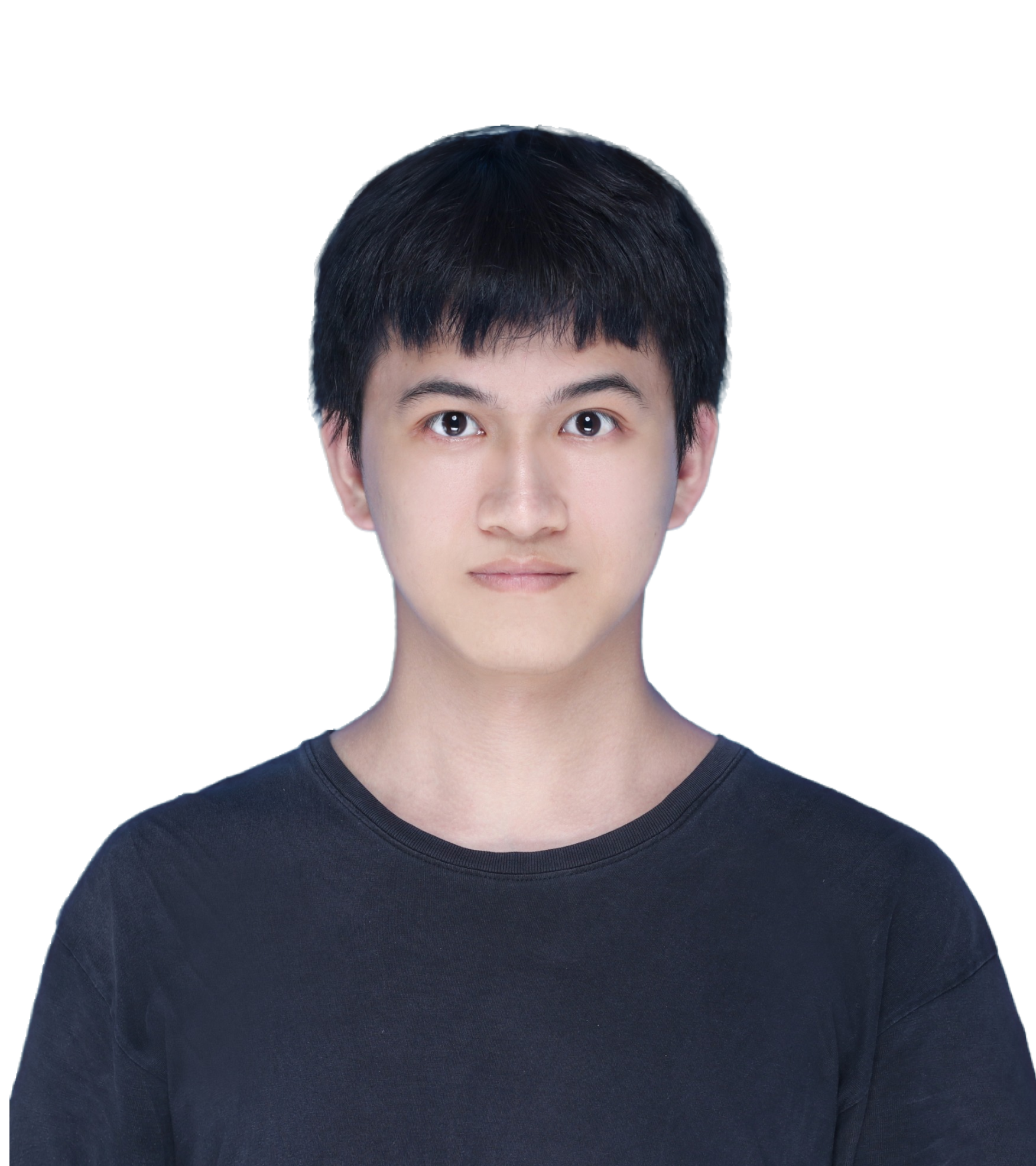}}]{Zhijie Qiu} received the B.E. degree in Software Engineering from the School of Software Engineering, South China University of Technology, China, in 2024. He is currently working toward the M.S. degree in the School of Software Engineering, South China University of Technology, China. His research interests include deep learning and model efficient inference.
\end{IEEEbiography}
\vspace{-25pt}

\begin{IEEEbiography}[{\includegraphics[width=1in,height=1.25in,clip,keepaspectratio]{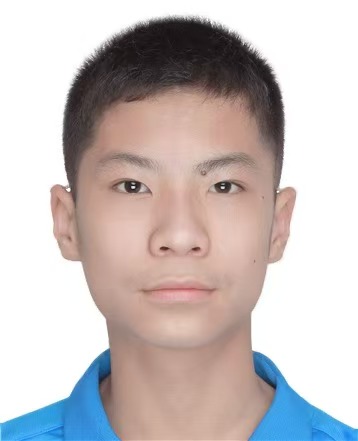}}]{Tingyu Wu}  is currently pursuing the B.E. degree in the School of Software Engineering, South China University of Technology, China. His research interests include large language model architectures and LLM-based agents.
\end{IEEEbiography}
\vspace{-25pt}

\begin{IEEEbiography}[{\includegraphics[width=1in,height=1.25in,clip,keepaspectratio]{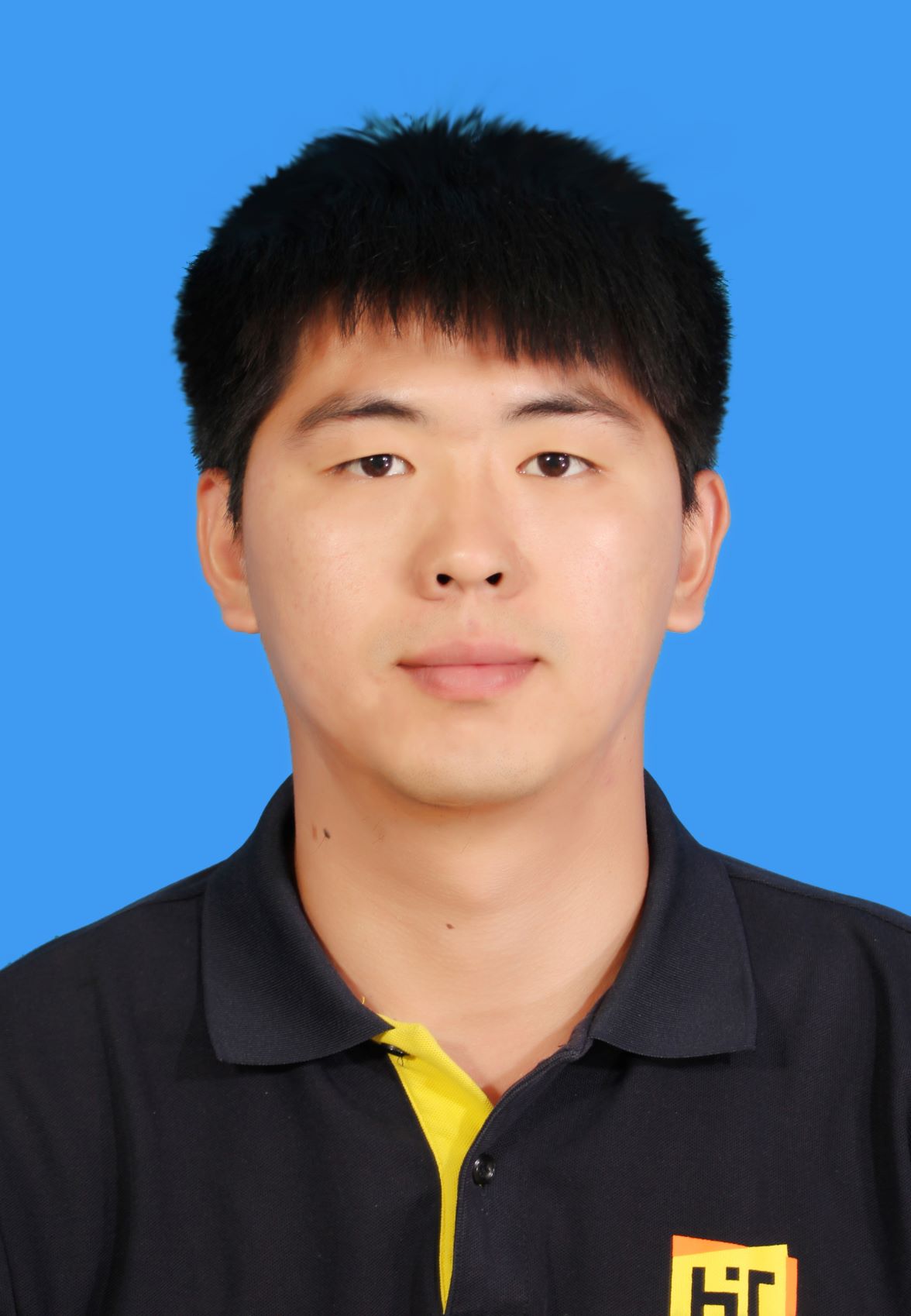}}]{Yingjian Li} received the B.Sc., M.Sc., and Ph.D. degrees from Harbin Institute of Technology, China, in 2016, 2018, and 2023, respectively. He was a visiting researcher of the Nanyang Technological
University (NTU).  Now he is an Assistant Research Fellow at Institute of Perceptual Intelligence, Pengcheng Laboratory, Shenzhen, China. He has published more than 20 journal (IEEE TIP, TMM, TAFFC, TCSVT, etc.) and conference (ACM-MM) papers.  His research interests include computer vision, multimodal leaning, and affective computing.
\end{IEEEbiography}
\vspace{-25pt}

\begin{IEEEbiography}[{\includegraphics[width=1in,height=1.25in,clip,keepaspectratio]{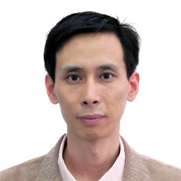}}]{Yaowei Wang}
(Member, IEEE) is a professor at the Harbin Institute of Technology, Shenzhen (HITSZ) and the Peng Cheng Laboratory. His research interests include artificial intelligence and large-scale video intelligent perception.
He has published over 100 papers in top-tier international journals and conferences and holds more than 50 authorized patents. He is a recipient of the "National High-Level Talent" honor. His work has been recognized with major awards, including the National Technological Invention Award (Second Prize), and both the Science and Technology Progress Award (First Prize) and the Technology Invention Award (First Prize) from the Chinese Institute of Electronics.
He serves as an Associate Editor for IEEE TCSVT and the chair of the IEEE Digital Retina System Working Group.
\end{IEEEbiography}
\vspace{-25pt}

\begin{IEEEbiography}[{\includegraphics[width=1in,height=1.25in,clip,keepaspectratio]{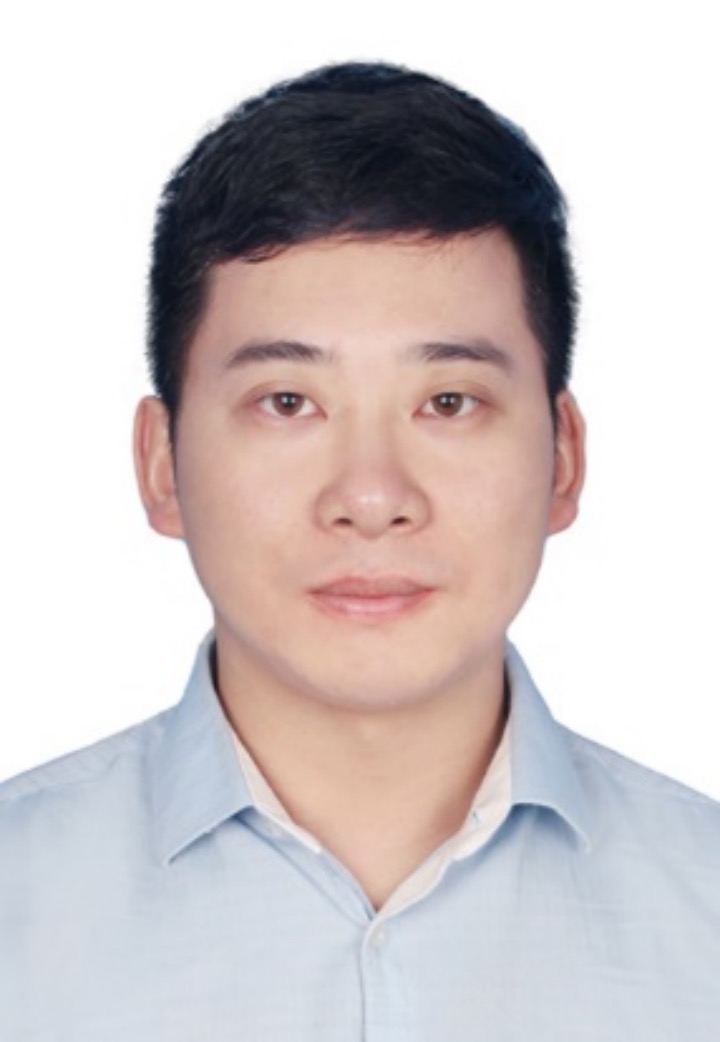}}]{Mingkui Tan}
is currently a professor with the School of Software Engineering at South China University of Technology. He received his Bachelor Degree in Environmental Science and Engineering in 2006 and Master degree in Control Science and Engineering in 2009, both from Hunan University. He received his Ph.D. degree in Computer Science from Nanyang Technological University, Singapore, in 2014. From 2014-2016, he worked as a Senior Research Associate on computer vision in the School of Computer Science, University of Adelaide, Australia. He also serves as an Associate Editor for IEEE Transactions on Pattern Analysis and Machine Intelligence. His research interests include machine learning, sparse analysis, deep learning, and large-scale optimization.
\end{IEEEbiography}
\vfill

\vfill

\end{document}


\markboth{Journal of \LaTeX\ Class Files,~Vol.~14, No.~8, August~2021}%
{Shell \MakeLowercase{\textit{et al.}}: A Sample Article Using IEEEtran.cls for IEEE Journals}

\IEEEpubid{0000--0000/00\$00.00~\copyright~2021 IEEE}

\appendices 
\section{Theoretical Analysis of Approximation Error}\label{app:theory}

A central question for any sparse attention mechanism is its capacity to accurately approximate the full attention output. In this section, we present a theoretical analysis that bounds the approximation error of \sexyname. We show that the error is bounded and can be explicitly controlled via the method's hyperparameters, particularly the threshold $\tau$ used in the offline configuration process.

For a single attention head, let the full attention output be $\mathbf{Att} = \text{softmax}(\mathbf{A}) \mathbf{V}$, where $\mathbf{A} = \frac{\mathbf{Q}\mathbf{K}^\top}{\sqrt{d_h}} \in \mathbb{R}^{L \times L}$ is the pre-softmax score matrix. \sexyname computes an approximation $\widetilde{\mathbf{Att}} = \text{softmax}(\widetilde{\mathbf{A}}) \widetilde{\mathbf{V}}$, where $\widetilde{\mathbf{A}} = \frac{\mathbf{Q}\widetilde{\mathbf{K}}^\top}{\sqrt{d_h}}$, and $\widetilde{\mathbf{K}}, \widetilde{\mathbf{V}} \in \mathbb{R}^{L \times d_h}$ contain only the selected global and local tokens (with other rows zeroed). The token selection is determined by the index set $\mathcal{S}_{\text{total}} = \mathcal{S} \cup \mathcal{S}_{\text{local}}$. 
We measure the error for a single query vector $\mathbf{q}_i$ (the $i$-th row of $\mathbf{Q}$) in terms of the $\ell_1$ norm between the full attention output vector and our approximation.

\begin{thm}[\textbf{Error Bound for a Single Query}]\label{thm:err}
Let $\gamma_i = \sum_{j \notin \mathcal{S}_{\text{total}}} \frac{\exp(\mathbf{A}_{ij})}{Z_i}$ be the total probability mass assigned by the full softmax to tokens not selected by \sexyname, where $Z_i$ is the normalization constant. Then the approximation error is bounded by
\begin{equation}
| \mathbf{Att}_i - \widetilde{\mathbf{Att}}_i |_1 \leq 2\gamma_i \cdot |\mathbf{V}|_\infty,
\end{equation}
where $\mathbf{Att}_i$ and $\widetilde{\mathbf{Att}}_i$ denote the $i$-th rows of the full and approximate attention outputs, respectively, $|\mathbf{V}|_\infty$ is the maximum absolute value in the value matrix.
\end{thm}

\begin{proof}

Let $\mathbf{s}_i = \text{softmax}(\mathbf{A}_i) = \frac{\exp(\mathbf{A}_i)}{Z_i}$ be the full attention probability distribution for query $i$.
Let $\widetilde{\mathbf{s}}_i = \text{softmax}(\widetilde{\mathbf{A}}_i)$ be the approximate distribution. Note that $\widetilde{\mathbf{s}}_i$ is defined only over the selected tokens; for mathematical convenience, we consider it a distribution over all $L$ tokens by setting $(\widetilde{\mathbf{s}}_i)_j = 0$ for $j \notin \mathcal{S}_{\text{total}}$.
Then, the attention outputs are:
$$\mathbf{Att}_i = \mathbf{s}_i \mathbf{V}, \quad \widetilde{\mathbf{Att}}_i = \widetilde{\mathbf{s}}_i \widetilde{\mathbf{V}}$$
Note that $\widetilde{\mathbf{V}}$ is $\mathbf{V}$ with rows zeroed out for unselected tokens, so $\widetilde{\mathbf{V}} = \mathbf{M} \odot \mathbf{V}$, where $\mathbf{M}$ is a masking matrix.
Thus, the error can be split using the triangle inequality:

\begin{equation}\label{eqn: E_i}
\begin{split}
    E_i &= | \mathbf{Att}_i - \widetilde{\mathbf{Att}}_i |_1\\
    &= \| \mathbf{s}_i \mathbf{V} - \widetilde{\mathbf{s}}_i \widetilde{\mathbf{V}} \|_1 \\
    &\leq \underbrace{\| (\mathbf{s}_i - \widetilde{\mathbf{s}}_i) \mathbf{V} \|_1}_{\text{Term I: Softmax Error}} + \underbrace{\| \widetilde{\mathbf{s}}_i (\mathbf{V} - \widetilde{\mathbf{V}}) \|_1}_{\text{Term II: Value Error}}.
\end{split}
\end{equation}

\textbf{1) Bounding Term II (Value Error)}: Since $\widetilde{\mathbf{V}}$ and $\mathbf{V}$ are identical for all selected tokens $j \in \mathcal{S}_{\text{total}}$, and $\widetilde{\mathbf{s}}_i$ is zero for unselected tokens, this term is zero.
\begin{equation}\label{eqn: termII}
    \text{Term II} = \| \widetilde{\mathbf{s}}_i (\mathbf{V} - \widetilde{\mathbf{V}}) \|_1 = 0.
\end{equation}

\textbf{2) Bounding Term I (Softmax Error)}: We now bound the difference between the two distributions $\mathbf{s}_i$ and $\widetilde{\mathbf{s}}_i$.
\begin{equation}
    \label{eqn: termII_o}
    \text{Term I} = \| (\mathbf{s}_i - \widetilde{\mathbf{s}}_i) \mathbf{V} \|_1 \leq \| \mathbf{s}_i - \widetilde{\mathbf{s}}_i \|_1 \cdot \|\mathbf{V}\|_\infty.
\end{equation}
We directly analyze the total variation. Let $Z_i = \sum_j \exp(A_{ij})$ and $\widetilde{Z}_i = \sum_{j \in \mathcal{S}_{\text{total}}} \exp(A_{ij})$. The difference is:
$$\| \mathbf{s}_i - \widetilde{\mathbf{s}}_i \|_1 = \sum_{j \in \mathcal{S}_{\text{total}}} \left| \frac{\exp(A_{ij})}{Z_i} - \frac{\exp(A_{ij})}{\widetilde{Z}_i} \right| + \sum_{j \notin \mathcal{S}_{\text{total}}} \frac{\exp(A_{ij})}{Z_i}
$$
The second term is precisely $\gamma_i$ by definition. The first term can be simplified:
$$
\sum_{j \in \mathcal{S}_{\text{total}}} \exp(A_{ij}) \left| \frac{1}{Z_i} - \frac{1}{\widetilde{Z}_i} \right| = \widetilde{Z}_i \cdot \frac{|Z_i - \widetilde{Z}_i|}{Z_i \widetilde{Z}_i} = \frac{|Z_i - \widetilde{Z}_i|}{Z_i}
$$
Recall that $\gamma_i = \sum_{j \notin \mathcal{S}_{\text{total}}} \frac{\exp(A_{ij})}{Z_i} = \frac{Z_i - \widetilde{Z}_i}{Z_i}$. Thus, we have
\begin{equation*}
\begin{aligned}
    \| \mathbf{s}_i - \widetilde{\mathbf{s}}_i \|_1 
    &= \sum_{j \in \mathcal{S}_{\text{total}}} \left| \frac{\exp(A_{ij})}{Z_i} - \frac{\exp(A_{ij})}{\widetilde{Z}_i} \right| + \gamma_i \\
    &= \gamma_i \left( \frac{\widetilde{Z}_i}{Z_i} + 1 \right) = \gamma_i \left( (1 - \gamma_i) + 1 \right) = \gamma_i (2 - \gamma_i) \leq 2\gamma_i.
\end{aligned}
\end{equation*}
Since $\gamma_i \in [0, 1]$, the term $\gamma_i (2 - \gamma_i)$ is always $\leq 2\gamma_i$. This is a tighter bound. Therefore, we get
\begin{equation}\label{eqn: Term I_gamma}
    \text{Term I} \leq 2 \gamma_i \cdot \|\mathbf{V}\|_\infty.
\end{equation}
According to the bounds for Term I in Eqn. (\ref{eqn: Term I_gamma}) and Term II in Eqn. (\ref{eqn: termII}), we obtain the result:
\begin{equation}\label{eqn: Ei_2gamma}
    E_i \leq 2 \gamma_i \cdot \|\mathbf{V}\|_\infty.
\end{equation}
\end{proof}

\textbf{Interpretation and Discussion}. Theorem~\ref{thm:err} provides a rigorous bound on the approximation error for each query position. The error scales linearly with $\gamma_i$, the total probability mass of the full attention is assigned to the tokens discarded by \sexyname.
\begin{itemize}[left=0pt]
    \item \textbf{Controllability:} Our offline calibration phase directly controls this error. By setting a threshold $\tau$ (e.g., 0.9) and ensuring that the aggregated attention mass of the selected tokens $\mathbf{a}_i \geq \tau$, we are effectively enforcing that $\gamma_i \leq 1 - \tau$ for the calibration data. This guarantees that the error bound is approximately limited to $ (1 - \tau) \cdot \|\mathbf{V}\|_\infty$ on the calibration distribution. By using a small calibration set, we ensure this bound generalizes to test inputs, which is demonstrated in Section V.
    \item \textbf{Role of Local Context:} The local window $\mathcal{S}_{\text{local}}$ ensures that the most recent tokens, which often have high attention scores for generative tasks, are always included. This prevents $\gamma_i$ from becoming large for these critical query positions, thus proactively minimizing the error bound.
    \item \textbf{Head-Specificity:} The error bound is per-head and per-query. This aligns with our design: heads with higher inherent redundancy (lower $\gamma_i$ for the same number of discarded tokens) can tolerate a sparser configuration ($\mathbf{p}^*$ with a lower budget) without violating the error constraint.
\end{itemize}
In conclusion, \sexyname is a principled approximation algorithm whose output deviation from full attention is theoretically bounded and explicitly managed by its configuration process. The empirical results showing maintained performance across diverse benchmarks (Section V) provide strong evidence that the values of $\gamma_i$ encountered in practice are indeed small, validating our theoretical analysis.

\begin{figure*}[h]
    \centering
    \includegraphics[width=\linewidth]{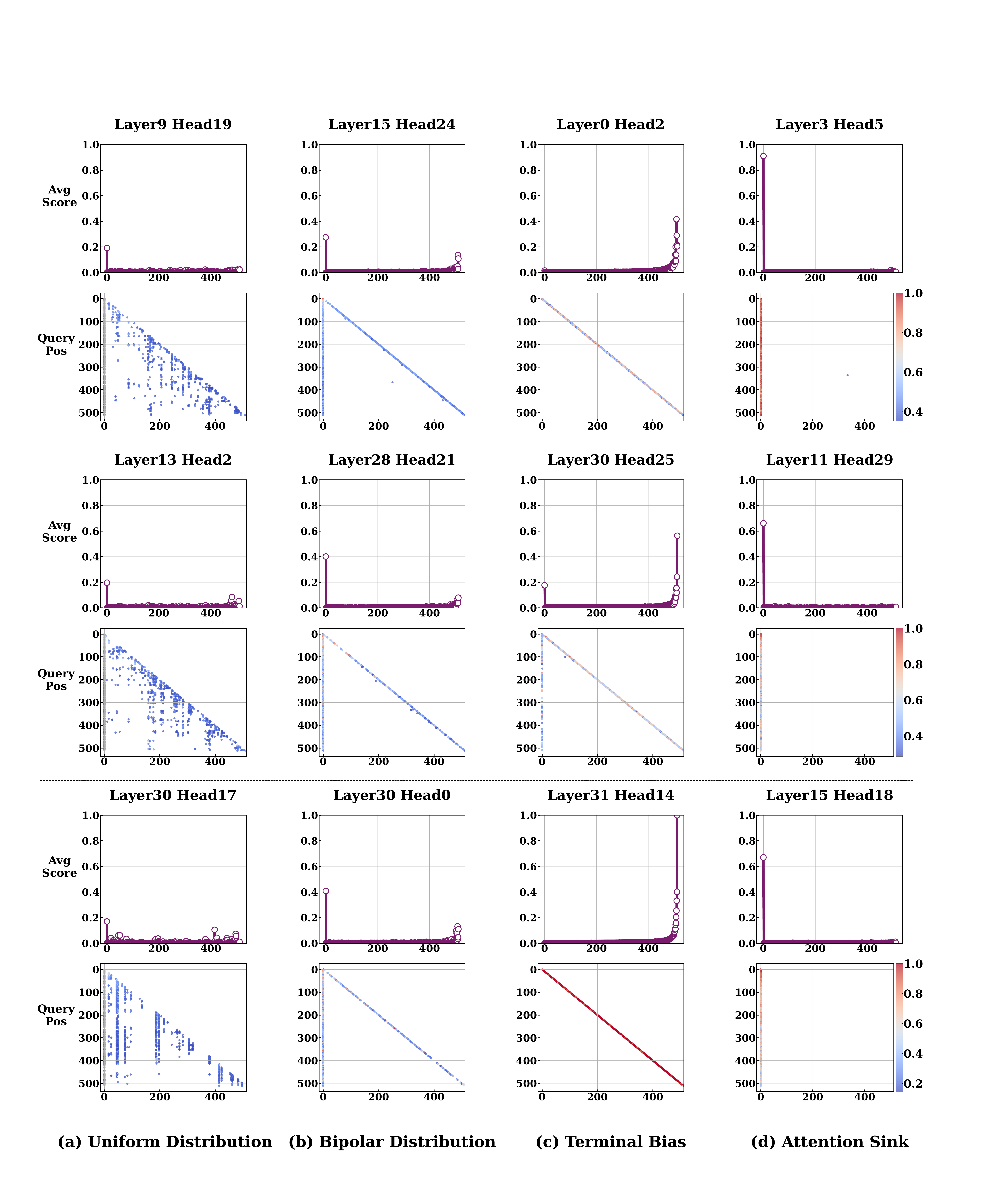}
    \caption{More visualization of attention score distributions across different layers and heads in LLaMA-3.1-8B-Instruct.}
    \label{fig:motivation_Llama}
\end{figure*}
\begin{figure*}[h]
    \centering
    \includegraphics[width=\linewidth]{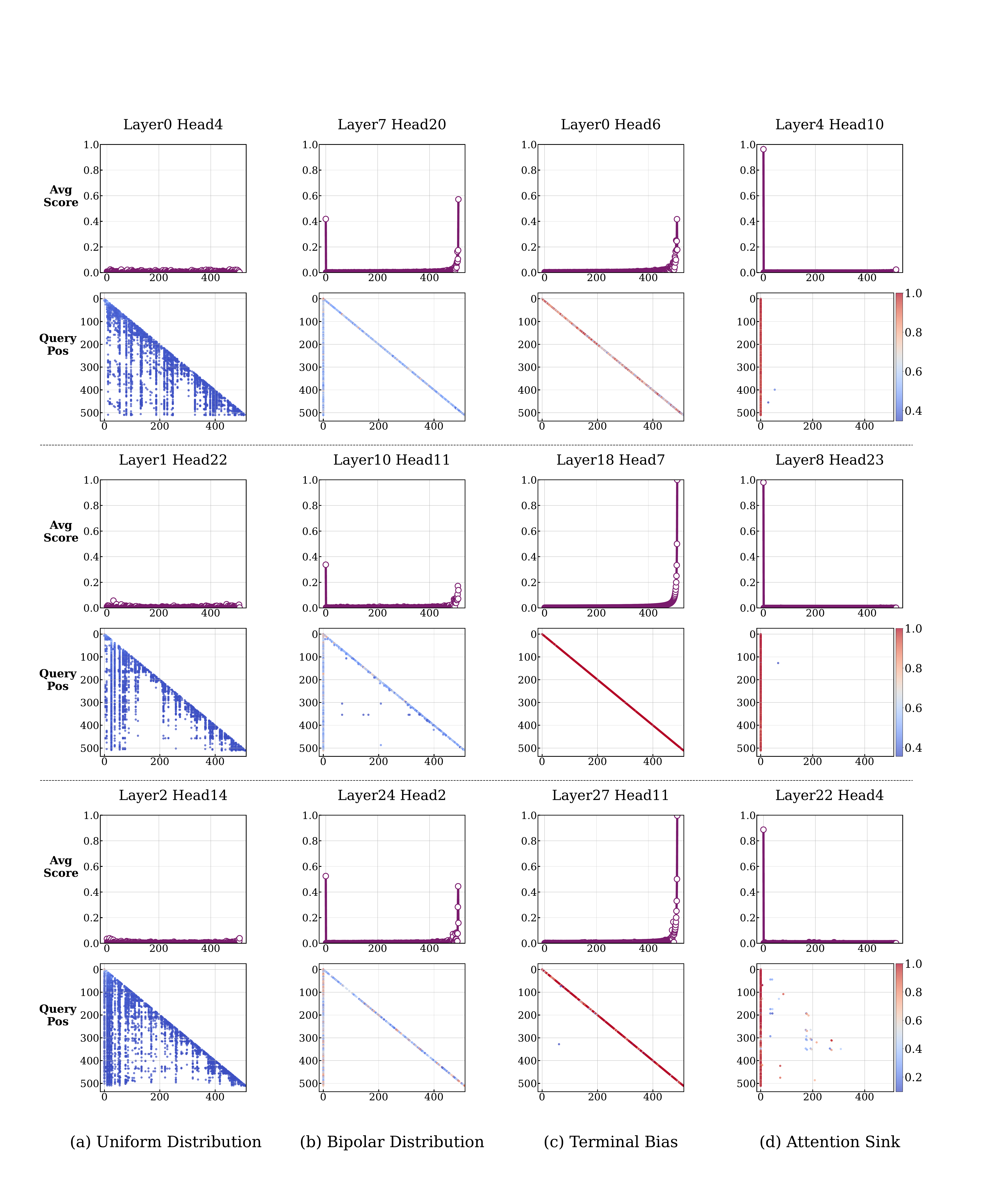}
    \caption{More visualization of attention score distributions across different layers and heads in Qwen2.5-7B-Instruct. (continued from previous page)}
    \label{fig:motivation_Qwen}
\end{figure*}

\section{More Visualizations of Attention Score Distributions}\label{sec:more_visual}
In this section, we provide more visualizations of attention score distributions for both LLaMA-3.1-8B-Instruct (Figure~\ref{fig:motivation_Llama}) and Qwen2.5-7B-Instruct (Figure~\ref{fig:motivation_Qwen}). Our analysis reveals consistent patterns across models: 
1) significant variation in attention distributions across different layers and heads; 
2) position-dependent scoring patterns throughout the sequence length;
3) the emergence of all four distribution types identified in Section~3 (uniform, bipolar, terminal bias, attention sink, and sparse activation). These observations hold regardless of model architecture, strongly motivating our design of a unified efficient attention mechanism that can automatically adapt to these heterogeneous patterns while maintaining computational efficiency. The cross-model consistency of these findings suggests that our approach may generalize well to other transformer-based LLMs.

\section{More Implementation Details}\label{supp:sec:implementation_details}

\subsection{More Details on Dataset and Evaluation Metrics}\label{app:dataset}

\textbf{LongBench~\cite{bai2023longbench}} represents a cutting-edge benchmark architecture engineered for the systematic assessment of LLMs across three critical dimensions: bilingual (Chinese-English) competence, multitask generalization, and long-context semantic processing. Its cross-linguistic design enables rigorous comparative analysis of multilingual contextual encoding abilities in scenarios demanding comprehension of extended textual sequences exceeding standard input boundaries. Structured into six overarching task categories—encompassing single-document question answering, multi-document reasoning, summarization, few-shot learning paradigms, synthetic linguistic tasks, and code completion workflows—the benchmark instantiates 21 meticulously designed subtasks that span the core application domains of long-text processing. Specifically, the corpus includes 14 English-language tasks, 5 Chinese-language tasks, and 2 code-oriented evaluation modules, with median sequence lengths ranging from 5K to 15K tokens and a total of 4,750 instances.

\textbf{RULER~\cite{hsieh2024ruler}} is a next-generation synthetic benchmark based on the NIAH paradigm, designed to evaluate the long-context capabilities of language model through configurable task complexity and sequence length. It extends traditional "needle" concepts into a taxonomy of semantic entities, relational patterns, and structural anomalies, enabling adjustable needle density to assess hierarchical information processing. The framework includes 13 subtasks across four categories: 8 retrieval tasks testing exact/semantic retrieval under noise, 3 multi-hop tracing tasks assessing sequential reasoning, 1 aggregation task evaluating contextual integration, and 1 complex QA task simulating multi-step inference. Its synthetic data pipeline ensures precise control over context length and needle correlations, supporting rigorous ablation studies and performance analysis of sequential memory, structural understanding, and multi-step reasoning.

\textbf{Massive Multitask Language Understanding (MMLU)~\cite{hendrycks2021measuring}} dataset is a comprehensive, multi-domain benchmark designed to rigorously assess the intellectual breadth and reasoning capabilities of language models across 57 diverse subject areas. These subjects span a wide spectrum of disciplines, including STEM, humanities, social sciences, and professional fields such as law and medicine, challenging models to demonstrate both factual knowledge and higher-order cognitive skills like analysis, inference, and application. Each question is crafted as a multiple-choice problem that requires nuanced understanding and logical reasoning, often surpassing surface-level pattern matching to evaluate a model’s ability to generalize knowledge in zero-shot and few-shot settings. By incorporating content ranging from elementary concepts to advanced expertise, MMLU simulates real-world scenarios where models must rely on their pretraining knowledge without task-specific fine-tuning. This interdisciplinary design makes MMLU a cornerstone for evaluating the adaptability, knowledge retention, and cross-domain generalization of modern language models, offering critical insights into their performance in realistic, knowledge-intensive environments.

\textbf{GSM8K~\cite{cobbe2021gsm8k}} is a benchmark designed to evaluate language models’ numerical reasoning and problem-solving abilities through grade-school math problems, consisting of 7,473 training and 1,319 test samples that focus on multi-step reasoning tasks involving 2 to 8 sequential steps, requiring models to break down complex problems into logical sub-tasks while demonstrating both conceptual understanding of mathematical principles like arithmetic, proportions, and algebra and precise computation skills. The dataset’s handcrafted problems exhibit rich linguistic variability with diverse narrative structures and real-world contextualizations, ensuring that models interpret nuanced language alongside solving equations, and it emphasizes chain-of-thought reasoning by encouraging models to articulate intermediate steps rather than provide direct answers, offering deeper insights into their logical processes.

\textbf{HumanEval~\cite{chen2021evaluating}} is a carefully curated code generation benchmark that consists of 164 hand-crafted Python programming problems, each designed to rigorously assess the coding capabilities of language models. Each problem in the dataset includes detailed specifications such as function signatures, docstrings, and a set of test cases, providing a comprehensive framework to evaluate both the syntactic correctness and functional accuracy of generated code. By covering a wide range of programming tasks—from basic algorithmic challenges to more complex logic-based problems—HumanEval ensures a thorough assessment of a model's ability to produce high-quality, executable code. The inclusion of unit tests further allows for precise validation of functional correctness, simulating real-world software development scenarios. As a result, HumanEval has become a cornerstone for evaluating the coding proficiency, logical reasoning, and problem-solving abilities of large language models in programming contexts.

\textbf{OlympiadBench}~\cite{he2024olympiadbench} is a rigorously constructed Olympiad-level bilingual multimodal benchmark designed to evaluate scientific reasoning capabilities of artificial intelligence systems. Comprising 8,476 challenging mathematical and physics problems sourced from international and national Olympiad competitions and the Chinese College Entrance Examination (Gaokao), this benchmark features expert-annotated step-by-step solutions, bilingual support (English and Chinese). By establishing a high standard for evaluating scientific reasoning beyond conventional benchmarks, OlympiadBench serves as an essential resource for advancing research toward artificial general intelligence with expert-level scientific capabilities.

\textbf{MT-Bench-101}~\cite{bai2024mt} is a comprehensive benchmark specifically designed for fine-grained evaluation of large language models' capabilities in multi-turn dialogues. The dataset features a three-tier hierarchical ability taxonomy that systematically assesses conversational skills across 13 distinct tasks, organized under seven detailed abilities and three overarching dimensions: perceptivity (context memory, understanding, and interference resistance), adaptability (rephrasing and reflection capabilities), and interactivity (questioning abilities). Comprising 1,388 carefully curated multi-turn dialogues with 4,208 conversational turns across 30 diverse topics, MT-Bench-101 provides golden context histories to ensure consistent evaluation conditions. Each dialogue was meticulously generated using task-specific prompts and rigorously validated by human annotators or advanced LLMs to guarantee high-quality assessment of multi-turn dialogue abilities beyond conventional single-turn benchmarks.

\begin{table*}[t]
\centering
\caption{Comparisons of different models on LongBench-E~\cite{bai2023longbench}.}
\label{tab:abl_sparsity}
\begin{tabular}{l|cccccc|c}
\hline
\multicolumn{1}{c|}{Methods} & Sin. Doc. QA & Mul. Doc. QA & Sum. & Few Shot & Syn. & Code & ~Average~ \\ \hline
Qwen2.5-7B-Instruct-128K & 48.75 & 52.24 & 27.81 & 65.00 & 52.00 & 61.14 & 51.16\\
~~$\bullet~$Random Sort & 47.19 & 50.78 & 26.63 & 63.35 & 51.50 & 60.07 & 49.92 \\
~~$\bullet~$Sort with Metric $\bh$  & 48.50 & 52.91 & 27.74 & 64.92 & 52.25 & 62.74 & \textbf{51.51}\\ 
\hline
\end{tabular}%
\end{table*}

\begin{table*}[t]
\centering
\caption{Performance on LongBench-E using Qwen2.5-7B-Instruct with different Concentration Indexes.}\label{tab:abl_centra_index}
\begin{tabular}{l|cccccc|c} 
\hline
\multicolumn{1}{c|}{Methods} & Sin. Doc. QA & Mul. Doc. QA & Sum. & Few Shot & Syn. & Code & ~Average~ \\ 
\hline
Entropy & 48.75 & 51.39 & 27.51 & 64.98 & 51.25 & 62.42 & 51.05 \\
Herfindahl-Hirschman Index & 48.50 & 52.91 & 27.74 & 64.92 & 52.25 & 62.74 & \textbf{51.51} \\
\hline
\end{tabular}
\end{table*}

\subsection{More Details on Parallel Implementation}\label{app:paralle_impl}
To address the challenges of efficient parallelization in \sexyname, we have developed specialized implementation strategies that overcome two critical barriers: non-contiguous memory access patterns and variable-length attention computation across attention heads.

\textbf{Memory Access Optimization}. The top-k token selection would lead to non-contiguous memory access, significantly degrading GPU performance. To mitigate this, we employ a parallel tokens gathering strategy that groups blocks where blocks within each group share the same top-k value. For each group, we perform the top-k selection in parallel across its blocks. Crucially, token selection is implemented by loading contiguous chunks of tokens within each block (since tokens are stored contiguously per block) and then gathering only the selected tokens. The selected tokens from all blocks are concatenated to form the global subsets $\mathbf{K}^G, \mathbf{V}^G$. This grouping strategy reduces memory fragmentation by minimizing random access scope, achieving near-optimal memory bandwidth utilization.

\textbf{Multi-head Parallelization}. Since each attention head requires a different number of tokens, this presents a significant challenge for parallelization. To address this, we flatten all head tokens into a single contiguous buffer and precompute offsets that map to each head's token region. During Triton kernel execution, each parallel thread uses these offsets to directly access the corresponding token region for its assigned head. This approach enables efficient parallel processing of variable-length multi-head attention computation without synchronization overhead, making full use of GPU parallelism while maintaining the head-aware design principle.

\textbf{Future Optimization Directions}. We plan to further enhance performance through latency-hiding scheduling techniques. Specifically, while computing attention between queries and the local subset $\mathbf{K}^L, \mathbf{V}^L$, we can concurrently perform global token selection in the background. Once local attention computation completes, the global subsets $\mathbf{K}^G, \mathbf{V}^G$ would be ready, allowing seamless transition to global attention without additional delay. While this approach shows promise for further reducing end-to-end latency, it requires sophisticated kernel-level scheduling and is left as future work.

\section{More Experimental Results}
\subsection{Effectiveness of Local-context Redundancy Metric}\label{sec:abl_h}
To validate the efficacy of our proposed \textit{Local-context Redundancy Metric} $\bh$ for redundancy estimation, we conduct an ablation study comparing our Online Core Context Selection strategy with a variant that randomly sorts blocks on LongBench-E with Qwen2.5-7B-Instruct. As shown in Table~\ref{tab:abl_sparsity}, our metric-driven approach achieves \textbf{1.59\%} average score improvement over the random baseline. This significant performance gap demonstrates that the proposed metric accurately quantifies block-wise redundancy levels and enables adaptive token retention requirements.

\subsection{Ablations on Concentration Index in the Redundancy Metric}
We conduct an ablation study to evaluate the effectiveness of the Herfindahl-Hirschman Index in Eq.6. As shown in Table~\ref{tab:abl_centra_index}, Replacing our Herfindahl-Hirschman Index in Eq.6 with a block-wise entropy measure results in a noticeable performance drop. The Herfindahl-Hirschman Index consistently delivers higher or comparable performance across most tasks, leading to a higher overall average score. This empirically validates that it is a more effective metric for identifying and preserving information-critical context.


\bibliography{tpami}
\bibliographystyle{IEEEtran}